\documentclass[10pt,journal,compsoc]{IEEEtran}

\usepackage{microtype}
\usepackage{graphicx}
\usepackage{makecell}
\usepackage{booktabs}
\usepackage{subfig}
\usepackage{hyperref}
\usepackage{amsmath}
\usepackage{algorithm}
\usepackage{algorithmic}
\usepackage{xcolor}

\newtheorem{assumption}{Assumption}
\newtheorem{theorem}{Theorem}

\def\abbr{LaNAS}

\begin{document}

\title{Sample-Efficient Neural Architecture Search by Learning Actions for Monte Carlo Tree Search}

\author{Linnan~Wang,
        Saining~Xie,
        Teng~Li,
        Rodrigo Fonseca,
        Yuandong Tian~\IEEEmembership{Member,~IEEE}
\IEEEcompsocitemizethanks{
\IEEEcompsocthanksitem L. Wang and R. Fonseca are with the Department
of Computer Science, Brown University, Providence, RI, 02905.\protect\\
\IEEEcompsocthanksitem S. Xie, T. Li, Y. Tian are with Facebook AI Research, Menlo Park, CA, 94025.
}
}

\IEEEtitleabstractindextext{
\begin{abstract}
Neural Architecture Search (NAS) has emerged as a promising technique for automatic neural network design. However, existing MCTS based NAS approaches often utilize manually designed action space, which is not directly related to the performance metric to be optimized (e.g., accuracy), leading to sample-inefficient explorations of architectures. To improve the sample efficiency, this paper proposes Latent Action Neural Architecture Search (LaNAS), which learns actions to recursively partition the search space into good or bad regions that contain networks with similar performance metrics. During the search phase, as different action sequences lead to regions with different performance, the search efficiency can be significantly improved by biasing towards the good regions. On three NAS tasks, empirical results demonstrate that LaNAS is at least an order more sample efficient than baseline methods including evolutionary algorithms, Bayesian optimizations, and random search. When applied in practice, both one-shot and regular LaNAS consistently outperform existing results. Particularly, LaNAS achieves 99.0\% accuracy on CIFAR-10 and 80.8\% top1 accuracy at 600 MFLOPS on ImageNet in only 800 samples, significantly outperforming AmoebaNet with $33\times$ fewer samples. Our code is publicly available at https://github.com/facebookresearch/LaMCTS.
\end{abstract}

}

\maketitle

\IEEEdisplaynontitleabstractindextext
\IEEEpeerreviewmaketitle

\IEEEraisesectionheading{\section{Introduction}\label{sec:introduction}}
\IEEEPARstart{D}{uring} the past two years, there has been a growing interest in Neural Architecture Search (NAS) that aims to automate the laborious process of designing neural networks. Starting from discrete state space and action space, NAS utilizes search techniques to explore the search space and find the best performing architectures concerning single or multiple objectives (\emph{e.g.}, accuracy, latency, or memory), and preferably with minimal search cost.

While it is impressive to find a good network architecture in a large search space, one component that is often overlooked is how to design the action space. Most previous methods use manually designed action space~\cite{zoph2016neural, baker2016designing, wang2019alphax}. Sec~\ref{sec:motivation} provides a simple example that different action spaces can significantly improve the search efficiency, by focusing on promising regions (e.g., deep networks rather than shallow ones) at the early stage of the search. Furthermore, compared to games that come up with a predefined action space (e.g., Atari or Go), learning action space is more suitable for NAS where the final network matters rather than specific action paths.

\begin{figure}[t]
  \begin{center}
    \includegraphics[height=6.0cm]{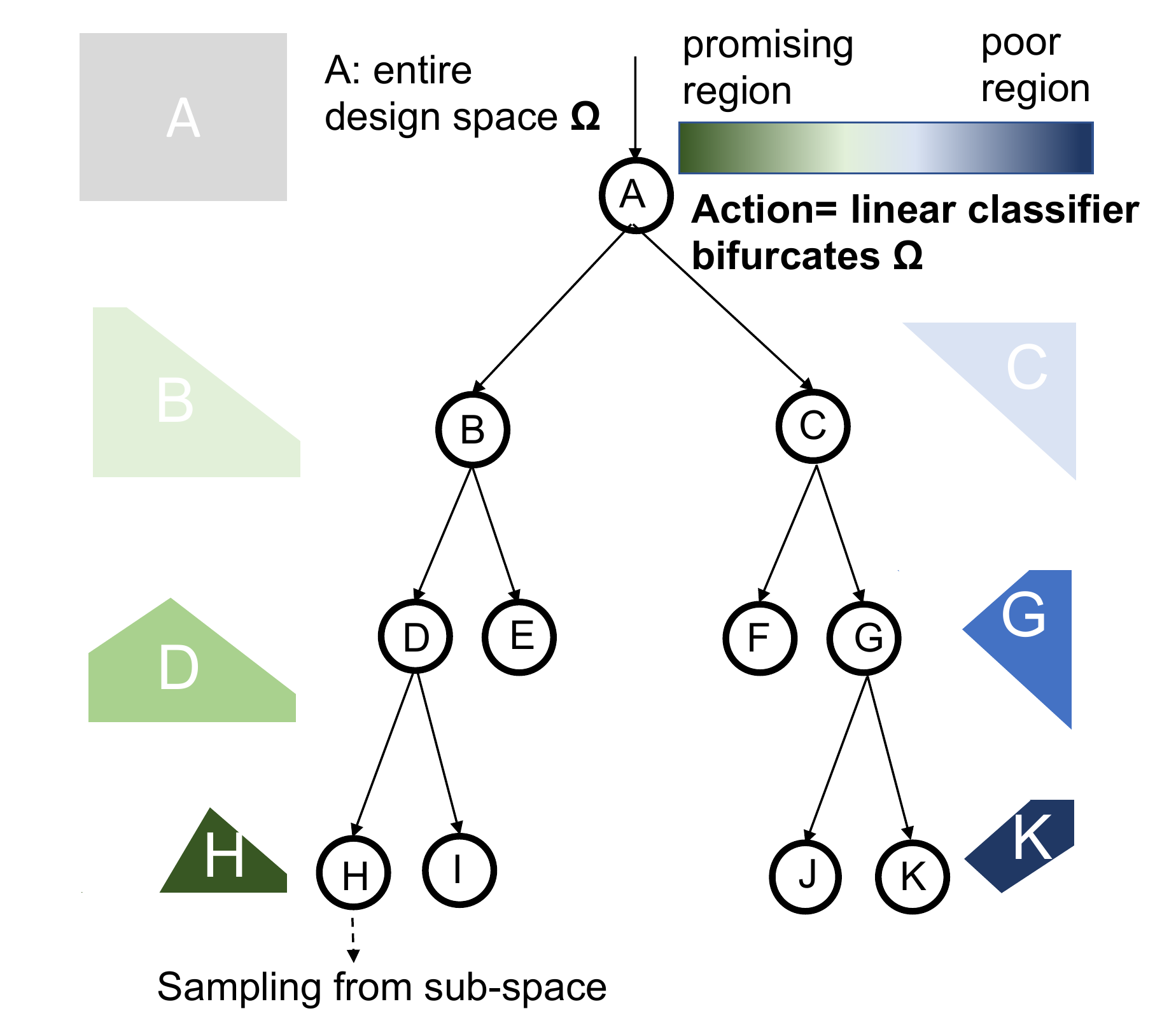}
    \end{center}
    \caption{Starting from the entire model space, at each search stage we learn an action (or a set of \emph{linear constraints}) to separate good from bad models for providing distinctive rewards for better searching. Fig.~\ref{partition-viz} in the appendix provides a visualization of the partitioning process in LaNAS.}
    \label{fig:teaser-fig}
\end{figure}

\begin{figure*}[t]
    \begin{center}
    \includegraphics[height=5.4cm]{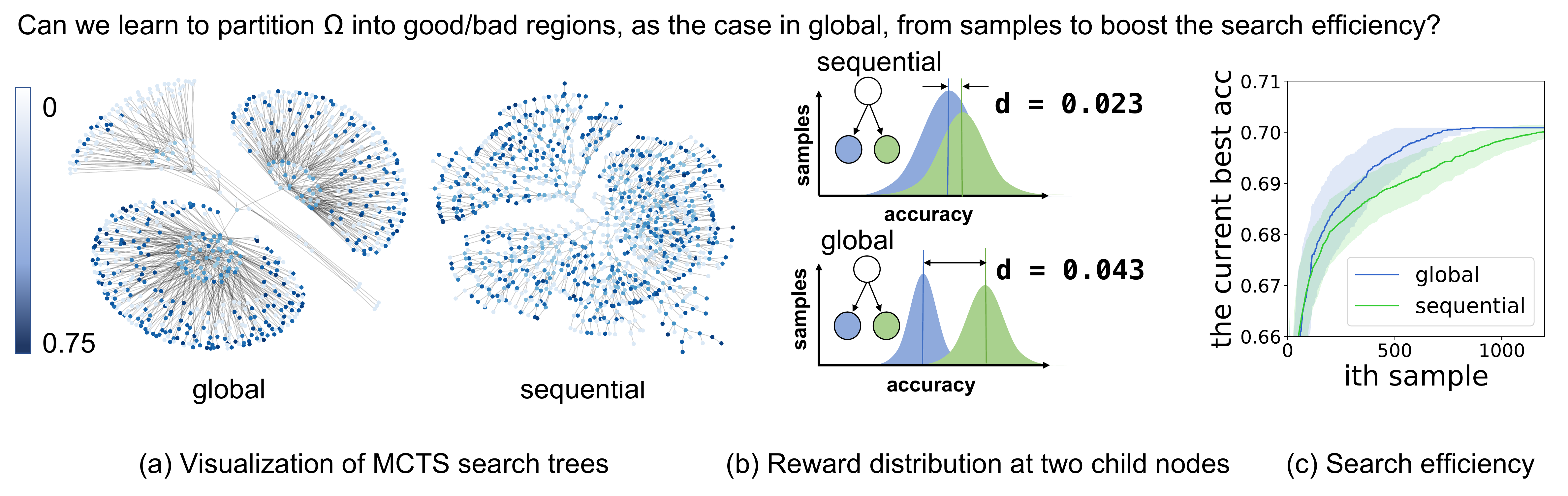}\label{fig:motivation}
    \end{center}
    \caption{\textbf{Illustration of motivation}: (a) visualizes the MCTS search trees using \texttt{sequential} and \texttt{global} action space. The node value (\emph{i.e.} accuracy) is higher if the color is darker.  (b) For a given node, the reward distributions for its children. $d$ is the average distance over all nodes. \texttt{global} better separates the search space by network quality and provides distinctive rewards in recognizing a promising path. (c) As a result, \texttt{global} finds the best network much faster than \texttt{sequential}. This motivates us to learn actions to partition the search space for the efficient architecture search.}
 \end{figure*}
 
Based on the above observations, we propose~\abbr~that learns \emph{latent actions} and prioritizes the search accordingly.  To achieve this goal, {\abbr} iterates between \emph{learning} and \emph{searching} stage. In the learning stage, {\abbr} models each action as a \emph{linear constraint} that bi-partitions the search space $\Omega$ into high-performing and low-performing regions. Such partitions can be done recursively, yielding a hierarchical tree structure, where some leaf nodes contain very promising regions, e.g. Fig.~\ref{fig:teaser-fig}. In the searching stage, LaNAS applies Monte Carlo Tree Search (MCTS) on the tree structure to sample architectures. The learned actions provide an abstraction of search space for MCTS to do an efficient search, while MCTS collects more data with adaptive exploration to progressively refine the learned actions for partitioning. The iterative process is jump-started by first collecting a few random samples.

Our empirical results show that LaNAS fulfills many desiderata proposed by~\cite{falkner2018bohb} as a practical solution to NAS: 1) \textit{\textbf{Strong final performance}}: we show that {\abbr} consistently yields the lowest regret on a diverse of NAS tasks using at least an order of fewer samples than baseline methods including MCTS, Bayesian optimizations, evolutionary algorithm, and random search. In practice, {\abbr} finds a network that achieves SOTA 99.0\% accuracy on CIFAR-10 and 80.8\% top1 accuracy (mobile setting) on ImageNet in only 800 samples, using 33$\times$ fewer samples and achieving higher accuracy than AmoebaNet~\cite{real2018regularized}. 2) \textit{\textbf{Use of parallel resources}}: LaNAS scales to, but not limited to, 500 GPUs in practice. 3) \textit{\textbf{Robustness \& Flexibility}}: The tree height and the exploration factor in UCB are only hyper-parameters in LaNAS, and we also conduct various ablation studies in together with a partition analysis to provide guidance in determining search hyper-parameters and deploying {\abbr} in practice.

LaNAS is an instantiation of Sequential Model-Based Optimization (SMBO)~\cite{hutter2009automated}, a framework that iterates between optimizing an acquisition function to find the next architecture to explore and obtaining the true performance of that proposed architecture to refine the acquisition.
Since evaluating a network is extremely expensive, SMBO is an efficient search framework for generalizing architecture performance based on collected samples to inform the search. Table.~\ref{baseline-methods} summarizes the attributes of existing search methods. Compared to Bayesian methods such as TPE, SMAC, and BOHB, LaNAS uses previous samples to learn latent actions, which converts complicated non-convex optimization of the acquisition functions into a simple traversal of hierarchical partition tree while still precisely captures the promising region for the sample proposal, therefore more efficient than Bayesian methods especially in high-dimensional tasks. We elaborate key differences to baseline methods (Table.~\ref{baseline-methods}) in Sec.~\ref{exp:alg-reasoning}.

\begin{table}[!tb]
  \setlength{\tabcolsep}{0.2em}
  \footnotesize
  \caption{Methods used in experiments and their attributes.}
  \label{baseline-methods}
  \begin{tabular}{l l l l l l}
    \toprule
        \textbf{Methods}   & \textbf{SMBO} & \begin{tabular}{@{}c@{}} \textbf{sampling} \\  \textbf{mechanism} \end{tabular} &
        \begin{tabular}{l} \textbf{ scalable to} \\ $|\Omega|\sim 10^{20}$ \end{tabular} &
        \begin{tabular}{c} \textbf{global}  \\ \textbf{search} \end{tabular} \\
        \midrule
        \begin{tabular}{l} HyperBand \cite{li2016hyperband}\end{tabular}& X &  \begin{tabular}{l} successive halving \end{tabular} & $\surd$ & X \\
        \begin{tabular}{l} BOHB \cite{falkner2018bohb}\end{tabular}& $\surd$ & non-convex optimization & X & X \\
        \begin{tabular}{l}SMAC  \cite{hutter2013evaluation}\end{tabular}& $\surd$  & non-convex optimization & X & $\surd$ \\
        \begin{tabular}{l} TPE \cite{bergstra2011algorithms}\end{tabular}& $\surd$ & non-convex optimization & X & $\surd$ \\
        \begin{tabular}{l} RE \cite{real2019regularized}\end{tabular}  & X & top-k random & $\surd$ & X \\
        \begin{tabular}{l} Random \cite{li2019random}\end{tabular}       & X & random & $\surd$ & $\surd$ \\
        \begin{tabular}{l} MCTS \cite{wang2018alphax}\end{tabular}       & X & UCB and search tree & $\surd$ & $\surd$ \\
        \begin{tabular}{l} LaNAS \end{tabular}   & $\surd$ & UCB and search tree & $\surd$ & $\surd$ \\
    \bottomrule
  \end{tabular}
  \\SMBO: Sequential Model Based Optimizations 
  \\$|\Omega|$ is the size of search space
\end{table}

\begin{figure*}[!t]
\begin{center}
\includegraphics[width=\textwidth]{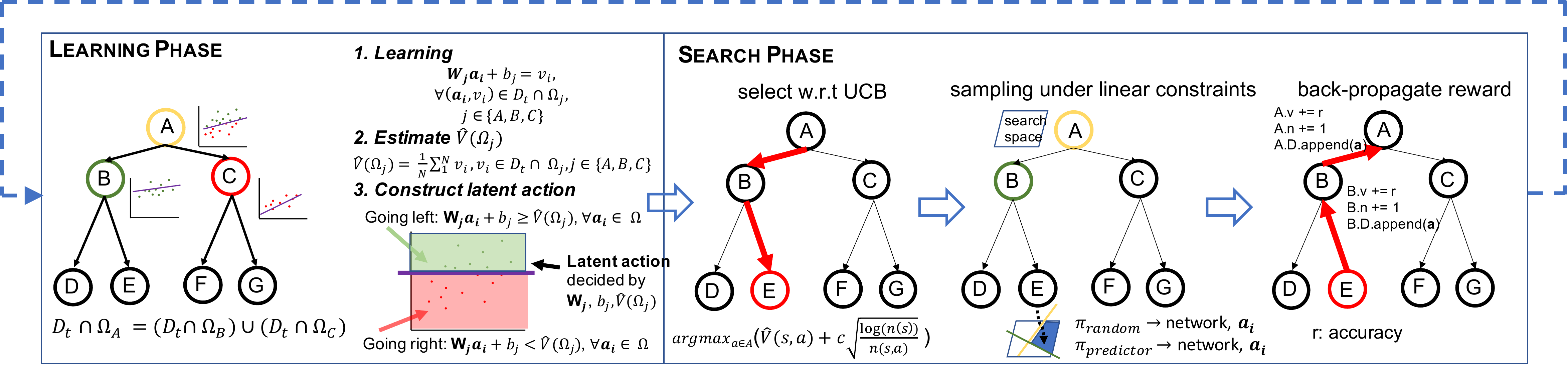} 
\caption{\textbf{An overview of LaNAS}: Each iteration of LaNAS comprises a search and learning phase. The search phase uses MCTS to sample networks, while the learning phase learns a linear model between network hyper-parameters and accuracies.}
\label{algorithm-overview}
\end{center}
\label{fig:perf-eval-datasets}
\end{figure*}

\section{A Motivating Example}
\label{sec:motivation}

To demonstrate the importance of action space in NAS, we start with a motivating example. Consider a simple scenario of designing a plain Convolutional Neural Network (CNN) for CIFAR-10 image classification. The primitive operation is a Conv-ReLU layer. Free structural parameters that can vary include network depth $L=\{1,2,3,4,5\}$, number of filter channels $C=\{32,64\}$ and kernel size $K=\{3\times 3, 5\times 5\}$. This configuration results in a search space of 1,364 networks. To perform the search, there are two natural choices of the action space: \texttt{sequential} and \texttt{global}. \texttt{sequential} comprises actions in the following order: {adding a layer $l$, setting kernel size $K_l$, setting filter channel $C_l$}. The actions are repeated $L$ times. On the other hand, \texttt{global} uses the following actions instead: \{Setting network depth $L$, setting kernel size $K_{1,\dots,L}$, setting filter channel $C_{1,\dots,L}$\}. For these two action spaces, MCTS is employed to perform the search. Note that both action spaces can cover the entire search space but have very different search trajectories. 

Fig.~\ref{fig:motivation}(a) visualizes the search for these two action spaces. Actions in \texttt{global} clearly separates desired and undesired network clusters, while actions in \texttt{sequential} lead to network clusters with a mixture of good or bad networks in terms of performance. As a result, the accuracy distribution of two branches (Fig.~\ref{fig:motivation}(b)) are separable for \texttt{global}, which is not the case for \texttt{sequential}. We also demonstrate the overall search performance in Fig.\ref{fig:motivation}(c) that \texttt{global} finds desired networks much faster than \texttt{sequential}.

This observation suggests that changing the action space can lead to very different search behavior and thus potentially better sample efficiency. In this case, an early exploration of network depth is critical. Increasing depth is an optimization direction that can potentially lead to better model accuracy. One might come across a natural question from this motivating example: is it possible to find a principle way to distinguish a good action space from a bad action space for MCTS? Is it possible to \emph{learn an action space} such that it can best fit the performance metric to be optimized?

\section{Learning Latent Actions for MCTS}

In this section, we describe {\abbr} that learns the action space for MCTS. Fig.~\ref{algorithm-overview} presents a high-level description of {\abbr}, of which the corresponding algorithms are further described in Alg.\ref{alg:lanas_procedures} in Appendix. The following describes a list of notations used in this paper.

\begin{align*}
    {a}_i&: \text{the ith sampled architecture} \\
    {v_i}&: \text{the performance metric of } \mathbf{a}_i \\
    {D_t}&: \text{the set of collected} (\mathbf{a}_i, v_i) \text{ at the search step t} \\
    {\Omega}&: \text{the entire search space}\\
    {\Omega_j}&: \text{the partition of $\Omega$ represented by the tree node $j$}\\
    {D_t \cap \Omega_j}&: \text{samples classified in } \Omega_j\\
    {V(\Omega_j)}&: \text{the mean performance metric in } \Omega_j \\
    {\hat{V}(\Omega_j)}&: \text{the estimated } V(\Omega_j) \text{ from } D_t \cap \Omega_j \\
    {f_j(\mathbf{a}_i)}&: \text{predicted performance by the regressor on node j} \\
    {n(s)}&: \text{the \#visits of tree node $s$} \\
    {v(s)}&: \text{the value  of tree node $s$} 
\end{align*}

\def\va{\mathbf{a}}

\subsection{Learning Phase}
\label{method-learningphase}
In the learning phase at iteration $t$, we have a dataset $D_t = \{(\va_i, v_i)\}$ obtained from previous explorations. Each data point $(\va_i, v_i)$ has two components: $\va_i$ represents an architecture in specific encoding (e.g., width=512 and depth=5, etc) and $v_i$ represents the performance metric estimated from training, or from pre-trained dataset such as NASBench-101, or estimated from a supernet in one-shot NAS.

\def\ch{\mathrm{ch}}

At any iteration, our goal is to learn a good action space from $D_t$ that splits $\Omega$ so that the performance of architectures is similar within each partition, but across partitions, the architecture performance can be easily ranked from low to high based on partitions. Fig.~\ref{fig:teaser-fig} shows this split can be recursively done to form a hierarchy; and our motivating example in Fig.~\ref{fig:motivation} suggests such partitions can help prioritize the search towards more promising regions, and improve the sample efficiency. In particular, we model the recursive splitting process as a tree. The root node corresponds to the entire model space $\Omega$, while each tree node $j$ corresponds to a region $\Omega_j$ (Fig.~\ref{fig:teaser-fig}). At each tree node $j$, we partition $\Omega_j$ into two disjoint regions $\Omega_j = \cup_{k\in(good, bad)} \Omega_k$, such that $\hat{V}(\Omega_{good}) > \hat{V}(\Omega_{bad})$ on each nodes. Therefore, a tree of these nodes recursively partitions the entire search space into different performance regions to achieve the target behavior in Fig.~\ref{fig:teaser-fig}. The following illustrates the algorithms in detail.

At each node $j$, we learn a regressor that embodies a latent action to split the model space $\Omega_j$. The linear regressor takes the portion of the dataset that falls into its own region $D_t \cap \Omega_j$, then the average performance of a region is estimated by
\begin{equation}
\hat{V}(\Omega_j) = \frac{1}{N}\sum_{v_i\in D_t\cap \Omega_j}v_i
\end{equation}

To partition $\Omega_j$ into $\Omega_{good}$ and $\Omega_{bad}$, we learn a linear regressor $f_j$
\begin{equation}
\underset{(\mathbf{a}_i, v_i)\in D_t\cap \Omega_j}{\text{minimize}} \sum(f_j(\va_i) - v_i)^2 
\end{equation}
Once learned, the parameters of $f_j$ and $\hat{V}(\Omega_j)$ form a linear constraint that bifurcates $\Omega_j$ into a good region ($ > \hat{V}(\Omega_j)$) and a bad region ($ \leq \hat{V}(\Omega_j)$). A visualization of this process is available in Fig.~\ref{algorithm-overview} (learning phase). For convenience, the left child always represents the good region. The partition threshold $\hat{V}(\Omega_j)$, combined with parameters of $f_j$, forms two latent actions at node $j$,
\begin{align*}
  \textbf{go-left}:& f_j(\va_i) > \hat{V}(\Omega_j) \\
  \textbf{go-right}:& f_j(\va_i) \leq \hat{V}(\Omega_j), \forall \va_i \in \Omega
\end{align*}

For simplicity, we use a full tree to initialize the search algorithm, leaving the tree height as a hyper-parameter. Fig.~\ref{fig:ablation_study} provides guidance in selecting the tree height.
Because the tree recursively splits $\Omega$, partitions represented by leaves follow $V(\Omega_{leftmost})>...>V(\Omega_{rightmost})$, with the leftmost leaf representing the most promising partition. Experiments in Sec.~\ref{analysis_search} validate the effectiveness of the proposed method in achieving the target behavior.

Note that we need to initialize each node classifier properly with a few random samples to establish an initial boundary in the search space. An ablation study on the number of samples for initialization is provided in Fig.~\ref{fig:ablation_study}(c). 
 
\subsection{Search Phase}
\label{lanas-search-phase}
Once actions are learned, the search phase follows. The search uses learned actions to sample more architectures $\va_i$, and get $v_i$ either via training or predicted from a supernet, then store ($\va_i, v_i$) in dataset $D_t$ to refine the action space in the next iteration. Note that during the search phase, the tree structure and the parameters of those classifiers are fixed and static. The search phase decides which region $\Omega_j$ on tree leaves to sample $\va_i$.

Given existing samples, a trivial go-left strategy, i.e. greedy-based search, can be used to exclusively exploit the most promising $\Omega_k$. However, the search space partitions or the latent actions learned from current samples can be sub-optimal such that the best model is located on any non-leftmost tree leaves. There can be good model regions that are hidden in the right (or bad) leaves that need to be explored.


To avoid this issue in a pure go-left search strategy, we integrate Monte Carlo Tree Search (MCTS) into the proposed search tree to enable adaptive explorations of different leaves. Besides, MCTS has shown great success in high dimensional tasks, such as Go~\cite{tian2019elf} and NAS~\cite{wang2018alphax}. MCTS avoids trapping into a local optimum by tracking both the number of visits and the average value on each node. For example, MCTS will choose the right node with a lower value if the left node with a higher value has been frequently visited before. More details are available in the paragraph of select w.r.t UCT below.

Like MCTS, our search phase also has \textit{select}, \textit{sampling} and \textit{back\-propagate} stages. {\abbr} skips the \textit{expansion} stage in regular MCTS since we use a static tree. At each iterations, previously sampled networks and their performance metrics in $D_t$ are reused and redirected to (maybe different) nodes, when initializing visitation counts $n(s)$ and node values $v(s)$ for the tree with updated action space. The details of these 3 steps are as follows.

1) \textbf{\textit{ select w.r.t UCB}}: UCB~\cite{auer2002finite} is defined by
\begin{equation}
	\label{ucb1_update}
	\pi_{UCB}(s) = \text{arg}\max_{a \in A} \left( \hat{V}(s, a) + c \sqrt{\frac{\log n(s)}{n(s, a)}} \right),
\end{equation}
$\pi_{UCB}$ chooses the action that yields the largest UCB score. In our case, the action is either going left or right on any non-leaf nodes. At the node $s$, $\hat{V}(s, a)$ represents the estimated value of the child node by taking action $a$, e.g. the value of left child by going left. $n(s)$ is the number of visits of the node $s$, which corresponds to the number of samples falling on the node represented partition. Similarly $n(s, a)$ represents the number of visits of the next node. Starting from $root$, we follow $\pi_{UCB}$ to traverse down to a leaf.

In $\pi_{UCB}$, the first term of $\hat{V}(s, a)$ represents the average value of next node after taking the action $a$ at the node $s$. By the construction of search tree, the average value of left child node is higher than the right. Therefore, $\pi_{UCB}$ degenerates to a pure go-left strategy if we set $c = 0$. The second term of $\log n(s)/n(s,a)$ represents the exploration, and $c$ is a hyper-parameter. $n(s)$ is same regardless of the action taken at node $s$, but the number of visits on the next node $n(s,a)$ can be drastically different. So a less visiting node with smaller $n(s,a)$ can increase $\pi_{UCB}$. Therefore, $\pi_{UCB}$ favors the node without any samples (dominated by the exploration term), or the node value is significantly higher (dominated by the exploitation term). Because of this mechanism, our algorithm can jump out of a sub-optimal action space learnt from current samples.

2) \textbf{\textit{sampling from a leaf}}: \textit{select} traverses a path from the root to a leaf, which defines a set of linear constrains for sampling. A node $j$ defines a constraint $l_j$ of $f_j(\va_i) \geq \hat{V}(\Omega_j), \forall \va_i \in \Omega$ if the path chooses the left child, and $f_j(\va_i) < \hat{V}(\Omega_j)$ otherwise. Therefore, the constraints from a path collectively enclose a partition $\Omega_j$ for proposing the new samples. Fig.~\ref{partition-viz} visualizes the process of partitioning along a search path.

Within a partition $\Omega_j$, a simple search policy is to use reject sampling: random sample until it satisfies the constraints. This is efficient thanks to limited numbers of constraints~\cite{gilks1995adaptive,hormann1995rejection,gorur2011concave}. Other strategies, e.g. Bayesian optimizations, can also be applied to sample from $\Omega_j$. Here we illustrate the implementation of both $\pi_{bayes}$ and $\pi_{random}$.

\begin{itemize}
	\item \textit{Random search based $\pi_{rand}$}: each component in the architecture encoding is assigned to a uniform random variable, and the vector of these random variables correspond to random architectures. For example, NASBench uses 21 Boolean variables for the adjacent matrix and 5 3-values categorical variables for the node list. The random generator uses 26 random integer variables, with 21 variables to be uniformly distributed in the set of [0, 1] indicating the existence of an edge, and 5 variables to be uniformly distributed in the set of [0, 1, 2] indicating layer types. $\pi_{rand}$ outputs a random architecture as long as it satisfies the path constraints.
	\item \textit{Bayesian search based $\pi_{bayes}$}: a typical Bayesian optimization search step consists of 2 parts: training a surrogate model using a Gaussian Process Regressor (GPR) on collected samples $D_t$; and proposing new samples by optimizing the acquisition function, such as Expected Improvement (EI) or Upper Confidence Bound (UCB). However, GPR is not scalable to large samples; and we used meta-DNN in~\cite{wang2019alphax} to replace GPR. Training the surrogate remains unchanged, but we only compute EI for architectures in the selected partition $\Omega_j$, and returns $\mathbf{a}_i$ with the maximum EI.
\end{itemize}

The comparisons of $\pi_{bayes}$ to $\pi_{random}$ is in Fig.~\ref{fig:sampleing-policy}. For the rest of the paper, we use $\pi_{random}$ in LaNAS for simplicity.

3) \textbf{\textit{ back-propagate reward}}: after evaluating the sampled network, {\abbr} back-propagates the reward, i.e. accuracy, to update the node statistics $n(s)$ and $v(s)$. It also back-propagates the sampled network so that every parent node $j$ keeps the network in $D_t \cap \Omega_j$ for training.

There are multiple ways to evaluate the performance of architecture, such as training from scratch, or predicted from a supernet as the case in one-shot NAS~\cite{pham2018efficient}. Among these methods, training every architecture from scratch (re-training) gives the most accurate $v_i$ but is extremely costly. While one-shot NAS is fairly cheap to execute as it only requires one-time training of a supernet for predicting $v_i$ for $\forall \mathbf{a}_i \in \Omega$, the predicted $v_i$ is quite inaccurate~\cite{sciuto2019evaluating}; therefore the architecture found by one-shot NAS is generally worse than the re-training approaches as indicated in Table.~\ref{open-domain-cifar}. In this paper, we try both one-shot based and training based evaluations. The integration of one-shot NAS is described as follows.

\begin{figure}[t]
  \begin{center}
    \includegraphics[width=0.95\columnwidth]{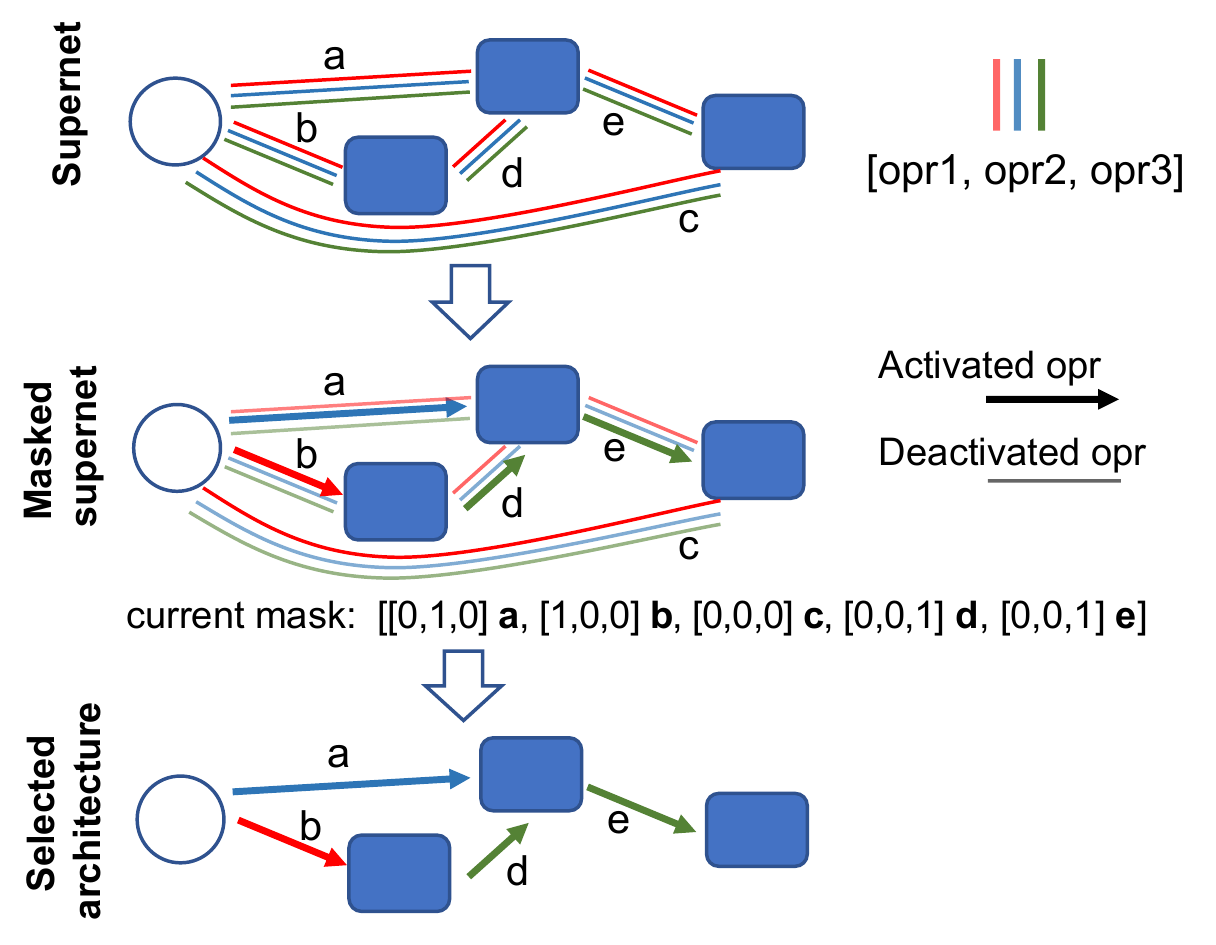}
  \end{center}
    \caption{\textbf{Integrating with one-shot NAS}: Before LaNAS comes into play, we pre-train a \emph{supernet} with a random mask at each iteration until it converges, i.e. decoupling the training and search so that we can benchmark different algorithms on the same supernet. During the search phase, the supernet remains static. When LaNAS evaluate a network $\va_i$, we transform the supernet to $\va_i$ by multiplying the mask corresponding to $\va_i$ as shown in the figure. \emph{Opr} stands for a layer type; we name edges from $a \rightarrow e$, and each edge can be one of the predefined layers or none. The figure shows there are 3 possibilities for a compounded edge, represented either by a 1x3 one-hot vector to choose a layer type to activate the edge, or a 1x3 zero vector to deactivate the edge.}
    \label{fig:supernet-fig}
\end{figure}

\begin{figure}[!t]
    \centering
     \subfloat[][supernet]{ \includegraphics[height=4.5cm, trim=5mm 0mm 0mm 0mm]{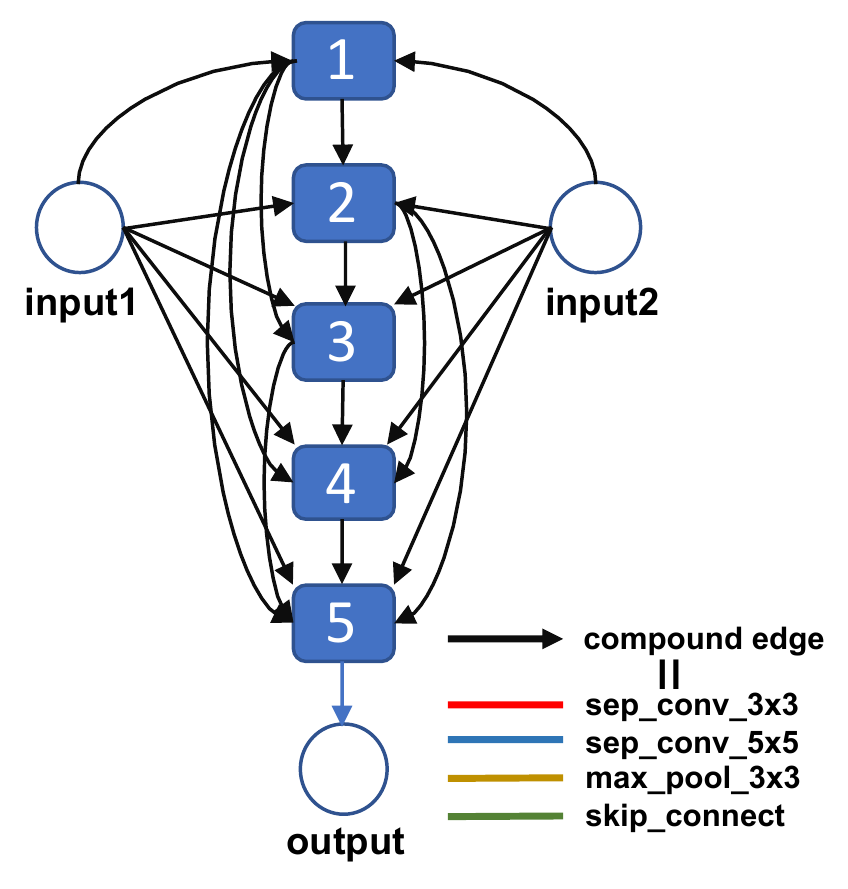}\label{supernet_nasnet_1}} 
 	 \subfloat[][masked supernet]{ \includegraphics[height=4.5cm, trim=5mm 0mm 0mm 0mm]{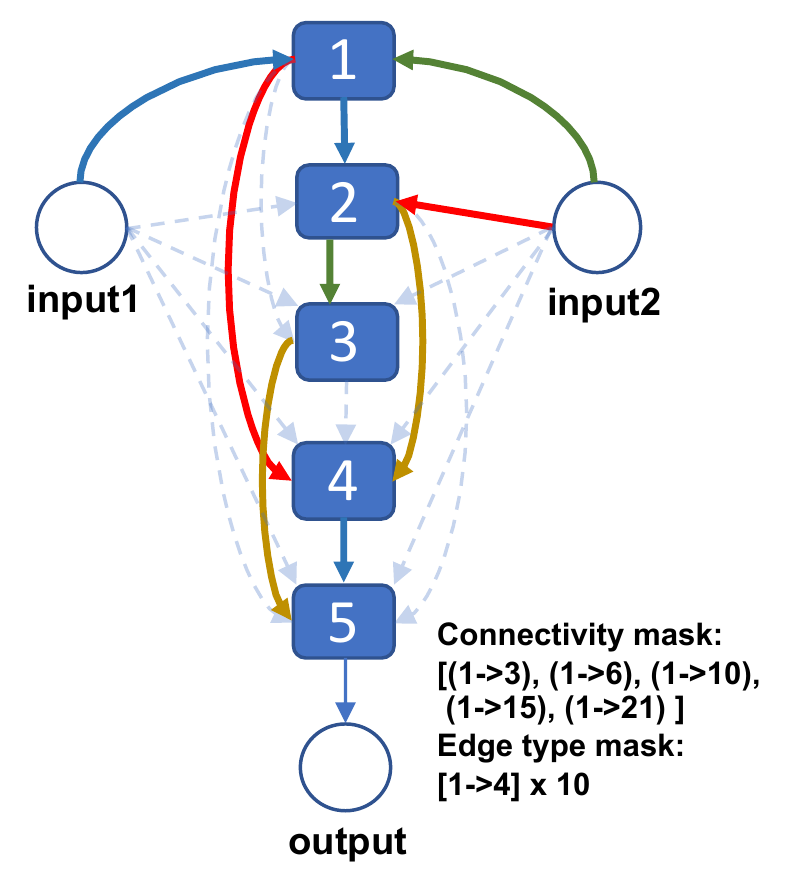}\label{supernet_nasnet_2}}
 	 \caption{\textbf{The cell structure of supernet} used in searching nasnet. The supernet structure of normal and reduction cell are same. (a) Each edge is a compound edge, consisting of 4 independent edges with the same input/output to represent 4 layer types. (b) Each node allows for two inputs from previous nodes. To specify a NASNet architecture, we use 5 variables for defining connectivity among nodes, and 10 variables for defining the layer type of every edge. Supernet can transform to any network in the search space by applying the mask.
 	 }\label{nasnet_supernet}
\end{figure}

\subsection{Integrating with one-shot NAS}
\label{oneshot-nas}

The key bottleneck in NAS is the expensive evaluation that trains a network from scratch. \cite{pham2018efficient} proposes a weight sharing scheme to avoid the model re-training using a supernet, which can transform to any architectures in the search space by deactivating extra edges to the network as shown in Fig.~\ref{fig:supernet-fig}. One popular approach for one-shot NAS is to formulate the training of supernet and the search on supernet as an integrated bi-level optimizations~\cite{liu2018darts}, while recent works~\cite{guo2019single, sciuto2019evaluating} show this also can be done by separating the training and the search. Our work primarily focuses on search efficiency, and we choose to separate the two procedures as it enables us to evaluate various algorithms on the same supernet (in Fig.~\ref{fig:perf-eval-datasets}(c)). We train the supernet by applying a random mask at each iteration following the same training pipeline/hyper-parameters in DARTs. After training the supernet, we fix the parameters of the supernet, then evaluate various search methods onto it. For example, LaNAS samples $\mathbf{a}_i$, masking the supernet to evaluate~$\mathbf{a}_i$ as illustrated by Fig.~\ref{fig:supernet-fig}; then $v_i$ is the validation accuracy evaluated from the masked supernet. The result ($\mathbf{a}_i, v_i$) is stored in $D_t$ to guide the future search. The following elaborates the design of the supernet for the NASNet search space, its training/search details, and the masking process.

\subsubsection{The design of supernet}
We have used two designs of supernet in this paper: one is for NASNet search space~\cite{zoph2018learning} to be evaluated on CIFAR10, and the other is for EfficientNet search space~\cite{tan2019efficientnet} to be evaluated on ImageNet.

\begin{itemize}
  \item \textit{Supernet for NASNet search space}: our supernet follows the design of NASNet search space~\cite{zoph2018learning}, the network of which is constructed by stacking multiple normal cells and reduction cells. Since the search space of normal/reduction cells is the same, the structure of supernet for both cells is also the same, shown in Fig.~\ref{supernet_nasnet_1}. The supernet consists of 5 nodes, and each node connects to all previous nodes. While a NASNet only takes 2 inputs, we enforce this logic by masking. Each edge consists of 4 independent edges that correspond to 4 types of layers.

  \item \textit{Supernet for EfficientNet search space}: we reused the supernet from~\cite{cai2019once}, please refer to Once-For-All for details.
\end{itemize}

\subsubsection{Transforming supernet to a specific architecture}
There are two steps to transform a supernet to a target architecture by masking. Here we illustrate it on the NASNet search space, and the procedures on EfficientNet are same. 

1) \textit{Specifying an architecture}: The NASNet search space specifies two inputs to a node, which can be any previous nodes. Therefore, we used 5 integers to specify the connections of 5 nodes, and each integer enumerates all the possible connections of a node. For example, node 4 in Fig.~\ref{supernet_nasnet_1} has 5 inputs, there are 5 possibilities $C(5,1)$ if two inputs are same, and 10 possibilities $C(5,2)$ for different inputs, adding up to 15 possible connections. Similarly, the possibilities for node1,2,3,5 are 3, 6, 10 and 21. Therefore, we use a vector of 5 integers with the range of $1\rightarrow3$, $1\rightarrow6$, $1\rightarrow10$, $1\rightarrow15$, and $1\rightarrow21$ to represent possible connections. After specifying the connectivity, we need to specify the layer type for each edge. In our experiments, a layer can be one of 3x3 separable convolution, 5x5 separable convolution, 3x3 max pooling and skip connect. Considering there are 10 edges in a NASNet cell, we use 10 integers ranging from 1 to 4 to represent the layer type chosen for an edge. Therefore, a NASNet can be fully specified with 15 integers (Fig.~\ref{supernet_nasnet_2}).

2) \textit{encoding to mask}: we need to change the encoding of 15 integers to a mask to deactivate the edges. Since the supernet in Fig~\ref{supernet_nasnet_1} has 20 edges, we use a 20x4 matrix, with each row as a vector to specify a layer or deactivation. The conversion is straightforward; if an edge is activated in the encoding, the edge is a one-hot vector, or a vector of 0s otherwise.

\subsubsection{Training supernet}
As explained in sec.~\ref{oneshot-nas}, we apply a random mask to each training iterations. We re-used the training pipeline from DARTs~\cite{liu2018darts}. To generate the random mask, we used 15 random integers (explained in generating random masks above) to generate a random architecture with their ranges specified in Fig.~\ref{supernet_nasnet_2}; then we transform the random encoding to a random mask, which is subsequently applied on supernet in training.

\subsubsection{Searching using pre-trained supernet}
 After training the supernet, it is fixed during search. A search method proposes an architecture $\va_i$ for the evaluation; we mask the supernet to $\va_i$ by following the steps in Fig.~\ref{fig:supernet-fig}, then evaluate the masked supernet to get $v_i$ for $\va_i$. The new $\va_i$, $v_i$ pair is stored in $D_t$ to refine the search decision in the next iteration. Since the evaluation of $\va_i$ is reduced to evaluating masked supernet on the validation dataset, this has greatly reduced the computation cost, enabling a search algorithm to sample thousands of $\va_i$ in a reasonable amount of time.

\subsection{Partition Analysis}
\label{partition-analysis}

The sampling efficiency is closely related to the partition quality of each tree node. Here we seek an upper bound for the number of samples in the leftmost leaf (the most promising region) to characterize the sample efficiency. LaNAS shows more speedup w.r.t random search as the size of the search space grows. Details are in sec~\ref{partition-analysis}.

\begin{assumption} 
 Given a search domain $\Omega$ containing finite samples $N$, there exists a probabilistic density $f$ such that $P( a < v < b ) = \int_{a}^{b}f(v)dv$, where $v$ is the performance of a network $\va$.
\end{assumption}

With this assumption, we can count the number of networks in the accuracy range of $[a, b]$ by $N*P(a \leq v \leq b)$. Since $v\in[0, 1]$ and the standard derivation $\sigma_v < \infty$, the following holds (\cite{median_mean})
\begin{equation}
	|E(\overline{v} -  M_{v}) | < \sigma_{v} 
\end{equation}
$\overline{v}$ is the mean performance in $\Omega$, and $M_{v}$ is the median performance. Note $v \in [0, 1]$, and let's denote $\epsilon=|\hat{v} - \overline{v}|$. Therefore, the maximal distance from $\hat{v}$ to $M_{v}$ is $\epsilon + \sigma_{v}$; and the number of networks falling between $\hat{v}$ and $M_{v}$ is $N*max( \int_{\hat{v}-\epsilon-\sigma_{v}}^{M_{v}}f(v)dv, \int^{\hat{v}+\epsilon+\sigma_{v}}_{M_{v}}f(v)dv )$, denoted as $\delta$. Therefore, the root partitions $\Omega$ into two sets that have $\leq \frac{N}{2} + \delta$ architectures.

\begin{theorem} Given a search tree of height = $h$, the sub-domain represented by the leftmost leaf contains at most $2*\delta_{max}(1-\frac{1}{2^h})+\frac{N}{2^h}$ architectures, and $\delta_{max}$ is the largest partition error from the node on the leftmost path.
\end{theorem}

The theorem indicates that LaNAS is approximating the global optimum at the speed of $N/2^h$, suggesting 1) the performance improvement will remain near plateau as $h\uparrow$ (verified by Fig~\ref{fig:ablation_study}(a)), while the computational costs ($2^h -1$ nodes) exponentially increase; 2) the performance improvement w.r.t random search (cost $\sim N/2$) is more obvious on a large search space (verified by Fig.5 (a)$\rightarrow$(c)).

\textit{Proof of Theorem}: In the worst scenario, the left child is always assigned with the large partition; and let's recursively apply this all the way down to the leftmost leaf $h$ times, resulting in $\delta^h + \frac{\delta^{h-1}}{2} + \frac{\delta^{h-2}}{2^2} + ... +\frac{N}{2^h} \leq$ $2*\delta_{max}(1-\frac{1}{2^h})+\frac{N}{2^h}$. $\delta$ is related to $\epsilon$ and $\sigma_{v}$; note $\delta \downarrow$ with more samples as $\epsilon \downarrow$, and $\sigma_{v}$ becomes more accurate.

\begin{figure*}[t]
\begin{center}
\includegraphics[width=0.9\textwidth]{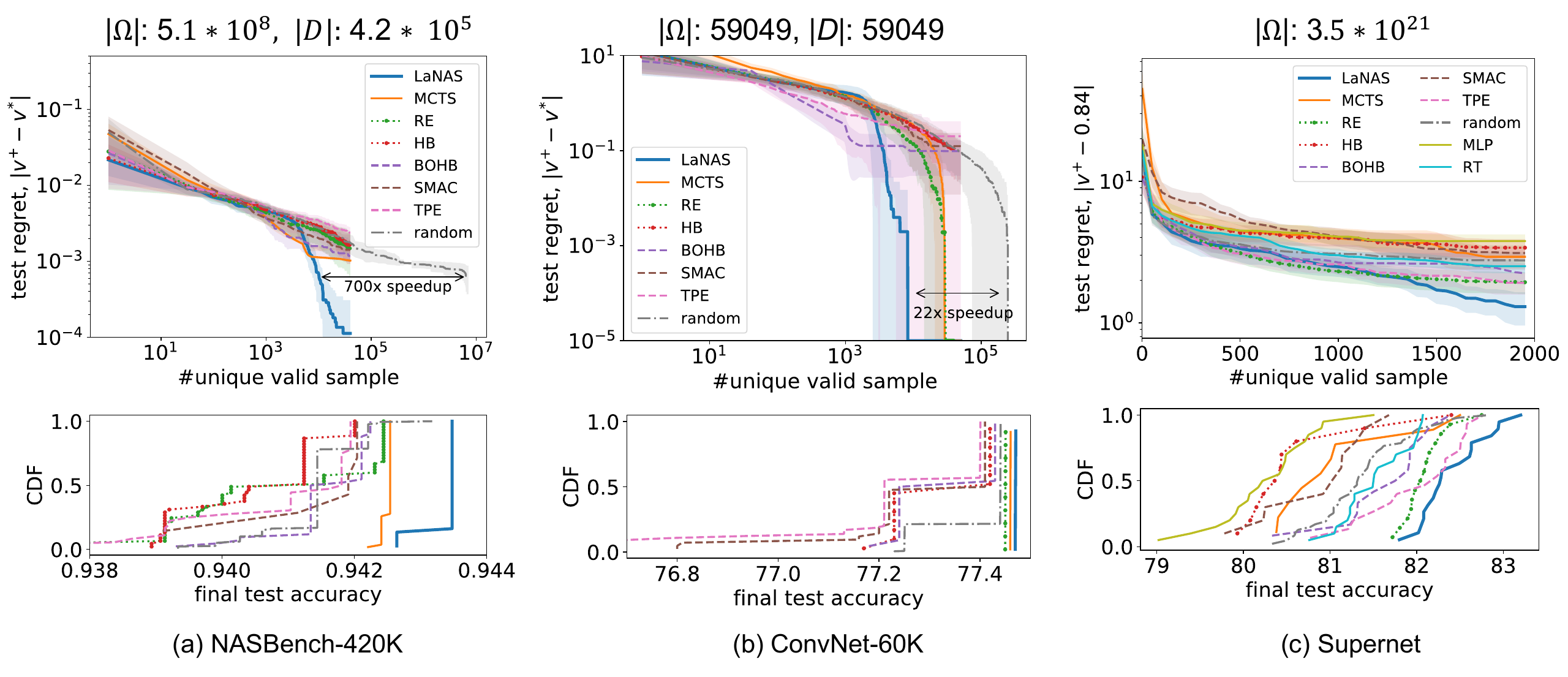}
\end{center}
\vspace{-0.14in}
\caption{The top row shows the time-course of test regrets of different methods (test regret between current best accuracy $v^{+}$ and the best in dataset $v^{*}$ with the interquartile range), while the bottom row illustrates Cumulative Distribution Function (CDF) of $v^+$ for each method at $4*10^4$ unique valid samples. ConvNet-60K compensates NASBench to test the case of $|D|=|\Omega|$, and supernet compensates for the case of $|\Omega|\gg|\Omega_{nasbench}|$, where $|D|, |\Omega|$ are the size of the dataset and search space, respectively. LaNAS consistently demonstrates the best performance in 3 cases.}
\label{fig:perf-eval-datasets}
\end{figure*}

\begin{figure*}[t]
\begin{center}
\includegraphics[width=\textwidth, trim=0mm 10mm 0mm 0mm]{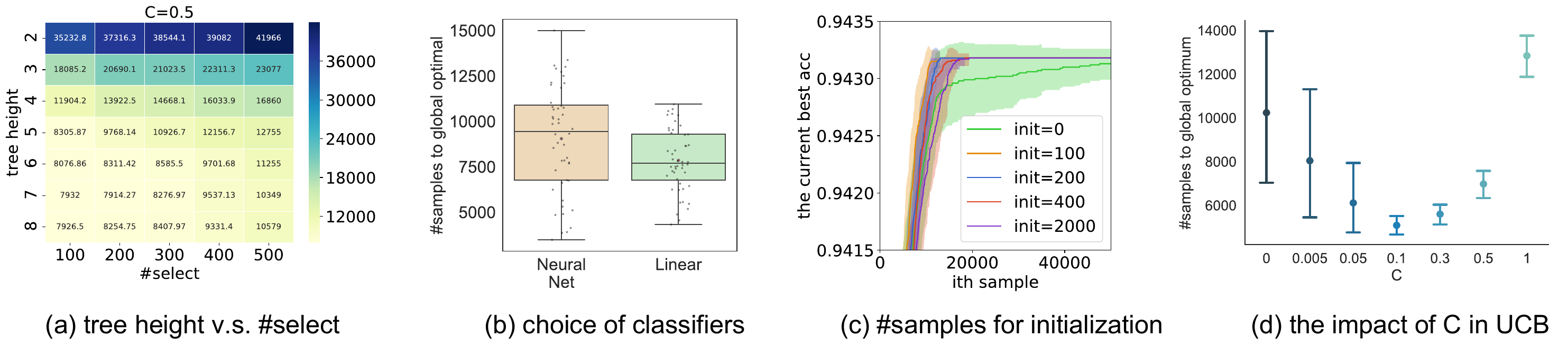}
\end{center}
\caption{\textbf{Ablation study}: (a) the effect of different tree heights and \#select in MCTS. The number in each entry is \#samples to reach global optimal. (b) the choice of predictor for splitting search space. (c) the effect of \#samples for initialization toward the search performance. (d) the effect of hyper-parameter $c$ in UCB on NASBench performance.} 
\label{fig:ablation_study}
\end{figure*}

\begin{figure*}[h]
\centering
\includegraphics[height=4cm]{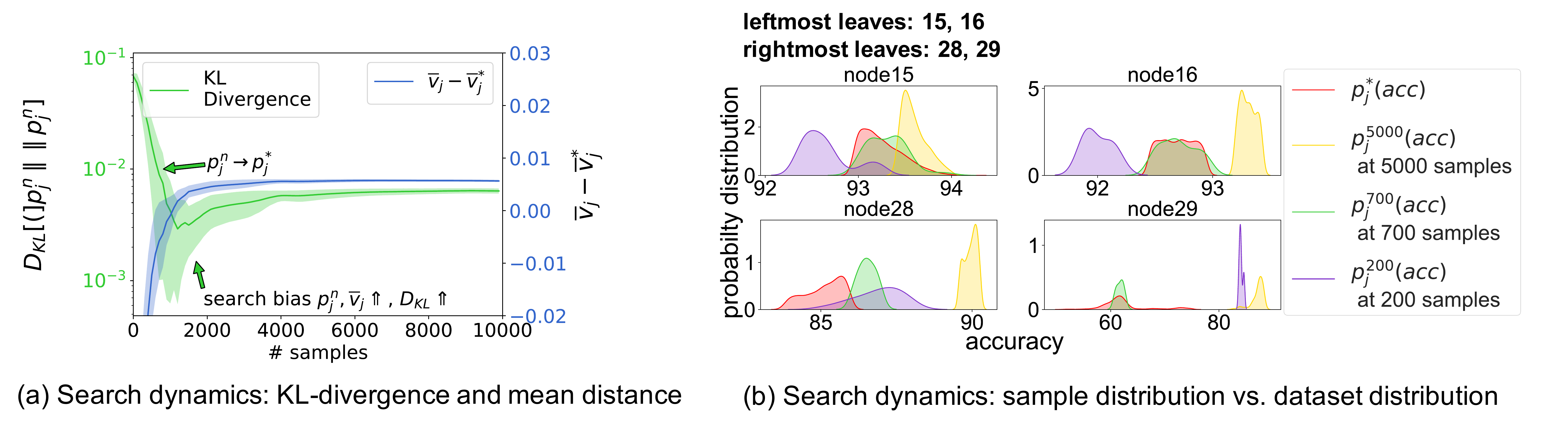}
\caption{\textbf{Evaluations of search dynamics}:(a) KL-divergence of $p_j$ and $p^*_j$ dips and bounces back. $\bar v-\bar v^*$ continues to grow, showing the average metric $\bar v$ over different nodes becomes higher when the search progresses. (b) sample distribution $p_j$ approximates dataset distribution $p^*_j$ when the number of samples $n \in [200, 700]$. The search algorithm then zooms into the promising sub-domain, as shown by the growth of $\bar v_j$ when $n \in [700, 5000]$. }
\label{fig:search-dynamics}
\end{figure*}

\section{Experiment}

\subsection{ Evaluating the search performance }
\label{exp:evaluations}

\subsubsection{Setup for benchmarks on NAS datasets}
\label{app:benchmark_dataset}

\textbf{Choice of NAS datasets/supernet}: NAS datasets record architecture-accuracy pairs for the fast retrieval by NAS algorithms to avoid time-consuming model retraining. This makes repeated runs of NAS experiments in a tractable amount of computing time to truly evaluate search algorithms. We use NASBench-101~\cite{ying2019bench} as one benchmark that contains over $4.2*10^5$ NASNet CNN models with edges $\leq$ 9 and nodes $\leq$ 7. To specify a network, search methods need 21 boolean variables for the adjacent matrix, and 5 3-value categorical variables for the node list\footnote{this is the best encoding scheme with the minimal missing architectures, which is also used in NASBench baselines.}, defining a search space of $|\Omega|=5*10^{8} \gg$ the size of dataset $|D|=4.2*10^5$. In practice, NASBench returns 0 for the missing architectures, which potentially introduces a bias in evaluations. Besides, NASBench is still several orders of smaller than a search space in practice, e.g. NASNet~\cite{zoph2018learning} $|\Omega|\sim 10^{20}$. To resolve these issues, we curate a ConvNet dataset having $5.9*10^{4}$ samples to cover the case of $|D| = |\Omega|$, and a supernet with $|\Omega| = 3.5*10^{21}$ to cover the case of $\Omega \gg \Omega_{nasbench}$ for benchmarks.

The curation of ConvNet-60K follows similar procedures in collecting 1,364 networks in sec.\ref{sec:motivation}, free structural parameters that can vary are: network depth $D=\{1,2,3,4,5,6,7,8,9,10\}$, number of filters $C=\{32, 64, 96\}$ and kernel size $K=\{3\times 3\}$, defining a $\Omega = 59049$. We train $\forall \va_i \in \Omega$ for 100 epochs, collect their final test accuracy $v_i$, store $(\mathbf{a}_i, v_i)$ in the dataset $D$. This small VGG-style, no residual connections, plain ConvNet search space can be fully specified with 10 3-value categorical variables, with each representing a type of filters. 

We use a supernet on NASNet search space, and sec.~\ref{oneshot-nas} provides the details about the curation and the usage of a supernet in evaluating the search efficiency.

\textbf{The architecture encoding}: 1) \textit{NASBench-101}: we used the architecture encoding of CIFAR-A in NASBench benchmarks from this repository\footnote{https://github.com/automl/nas\_benchmarks}, as the discrepancy between the size of the dataset and the search space is the minimal. Specifically, an architecture is encoded with 21 Boolean variables and 5 3-values categorical variables, with each value corresponding to 3 layer types, i.e. 3x3 convolution, 1x1 convolution, and max-pool. The 21 Boolean variables represent the adjacent matrix in NASBench, while the 5 categorical variables represent the nodelist in NASBench. Therefore, $|\Omega|=2^{21}*3^5 = 5.1*10^8$. 2) \textit{ConvNet-60K}: we used 10 3-values categorical variables to represent a VGG style CNN up to depth = 10, with each value correspond to 3 types of convolution layers , i.e. (filters=32, kernel = 3), (filters=64, kernel = 3) and (filters=96, kernel = 3). Therefore, $|\Omega|=59049$. 3) \textit{Supernet}: Since supernet implements the NASNet search space, the encoding of a supernet is same as the one used for NASBench-101. 

\textbf{Choice of baselines and setup for LaNAS}: we adopt the same baselines established by NASBench-101, and the same implementations from this public release\footnote{https://github.com/automl/nas\_benchmarks}. These baselines, summarized in Table.~\ref{baseline-methods}, cover diverse types of search algorithms. Regularized Evolution (RE) is a type of evolutionary algorithm that achieves SoTA performance for image recognition. While traditional BO method~\cite{snoek2012practical} suffers from the scalability issue (e.g. the computation cost scales $\mathcal{O}(n^3)$ with \#samples), random forest-based {Sequential Model-based Algorithmic Configuration} (SMAC) and {Tree of Parzen Estimators} (TPE) are two popular solutions by using a scalable surrogate model. HyperBand (HB) is a resource-aware (e.g. training iterations or time) search method, and {Bayesian optimization-based HyperBand} (BOHB) extended HB for strong any time performance. In addition to baselines in NASBench-101, we also added MCTS to validate latent actions. We have extensively discussed LaNAS v.s. these baselines in sec~\ref{exp:alg-reasoning}. 

In LaNAS, the height of the search tree is 8; we used 200 random samples for the initialization, and \#select = 50 (the number of samples from a selected partition, see sec.~\ref{exp:ablations} ).

\subsubsection{Details about Ensuring Fairness}
\label{fairness_setup}
1) \textit{The encoding scheme}: the encoding scheme decides the size of the search space, thereby significantly affecting the performance. We ensure LaNAS and MCTS to use the same encoding as NASBench baselines on both datasets.

2) \textit{Repeated samples}: we noticed NASBench baselines allow the same architecture to be sampled at different steps, and we modified LaNAS and MCTS to be consistent with baselines (by not removing samples from the search space).

3) \textit{Evaluation metric}: we choose the number of unique, valid\footnote{we define valid samples as the samples in NASBench, and invalid as those in the search space but not in NASBench.} samples instead of time to report the performance as model-free methods such as random search can easily iterate through the search space in a short time. 

4) \textit{Optimizing hyper-parameters}: the hyper-parameters of baselines are set w.r.t ablation studies in NASBench-101, and we also tuned LaNAS and MCTS for the benchmark.

5) \textit{Sufficient runs with different random seeds}: each method is repeated 100 runs with different random seeds.

\subsubsection{Empirical results}
The top row of Fig.~\ref{fig:perf-eval-datasets} shows the mean test regret, $|v^{+}-v^{*}|$ where $v^{+}$ is the current best and $v^{*}$ is the best in a dataset, along with the 25th and 75th percentile of each method through the course of searching, and the bottom row shows the robustness of methods at 40000 UVS on NASBench and ConvNet-60K, and 2000 UVS on supernet, respectively. Noted not all baselines are guaranteed to reach the global optimum, 40000 is the maximum UVS collected in 3 CPU days for all baselines on datasets, and 2000 is the maximum UVS on supernet in 3 GPU days.

We made the following observations: 1) \textit{strong final performance}: LaNAS consistently demonstrates the strongest final performance on 3 tasks. On NASBench (Fig.~\ref{fig:perf-eval-datasets}(a)), the final test error of LaNAS is 0.011\%, an order of smaller than the second best (0.137\%); Similarly, on supernet (Fig.~\ref{fig:perf-eval-datasets}(c)), the highest test accuracy found by LaNAS is 83.5\%, 0.75\% better than the second best.

2) \textit{good for one-shot NAS}: the strong final performance of LaNAS is more relevant to one-shot NAS as shown in Fig~\ref{fig:perf-eval-datasets}(c), as evaluations are fairly cheap.

3) \textit{performance behavior}: across 3 experiments, the performance of LaNAS is comparable to Random Search in the few hundreds of samples $\sim500$, and surpass baselines afterward. As an explanation for this behavior, we conduct a set of controlled experiments in Appendix.~\ref{analysis_search}. We conclude that LaNAS needs a few hundreds of samples to accurately estimate boundaries, thereby good performance afterward.

4) \textit{faster in larger $\Omega$}: LaNAS is 700x and 22x faster than \textit{random} in reaching similar regrets on NASBench-101 and ConvNet-60K. The empirical results validate our analysis (Appendix~\ref{partition-analysis}) that better performance w.r.t random search are observable on a larger search space.\\

\subsubsection{Discussions of baselines v.s. LaNAS}
\label{exp:alg-reasoning}
 Like existing SMBO methods, LaNAS uses a tree of linear regressor as surrogate $S$ in predicting the performance of unseen samples, and its $S$ is proven to be quite effective as the resulting partitions clearly separate good/bad $\Omega_j$ (validated by Fig.~\ref{fig:search-dynamics}(a), and Fig.~\ref{fig:search-dynamics}(b) in Appendix~\ref{analysis_search}). Besides, LaNAS uses $\pi_{ucb}$ in MCTS as the acquisition to trade-off between exploration and exploitation; All together makes LaNAS more efficient than non-SMBO baselines. For example, RS relies on blind search, leading to the worst performance. RE utilizes a static exploration strategy that maintains a pool of top-$K$ architectures for random mutations, not making full use of previous search experience. MCTS builds online models of both performance and visitation counts for adaptive exploration. However, without learning action space, the performance model at each node cannot be highly selective, leading to inefficient search (Fig.~\ref{fig:motivation}). The poor performance of HB attributes to the low-rank correlation between the performance at different budgets (Fig.7 in Supplement S2 of \cite{ying2019bench}).
 
\def\va{\mathbf{a}}
 
Compared to Bayesian methods, LaNAS learns the state partitions to simplify optimization of the acquisition function $\phi$. With learned actions, optimization is as simple as a quick traverse down the tree to arrive at the most performant region $\Omega_j$, regardless of the size of $|\Omega|$ and the dimensionality of tasks, and random sample a proposal within. Therefore, LaNAS gets a near-optimal solution to $\max_{\va_i\in \Omega}\phi(\va_i)$ but without explicit optimization. In contrast, Bayesian methods such as SMAC, TPE, and BOHB use iterated local search/evolutionary algorithm to propose a sample, which quickly becomes intractable on a high dimensional task, e.g. NAS with $|\Omega| > 10^{20}$. As a result, a sub-optimal solution to $\max_{\va_i\in\Omega} \phi(\va_i)$ leads to a sub-optimal sample proposal, thereby sub-optimal performance (shown in Fig~\ref{suboptimal_opt}, Appendix). Consistent with~\cite{wang2014bayesian}, our results in Fig.~\ref{fig:perf-eval-datasets} also confirms it. For example, Bayesian methods, BOHB in particular, perform quite well w.r.t LaNAS on ConvNet (the low dimensional task) especially in the beginning when LaNAS has not learned its partitions well, but their relative performances dwindle on NASBench and supernet (high dimensional tasks) as the dimensionality grows, $|\Omega_{supernet}| \gg |\Omega_{nasbench}| \gg |\Omega_{convnet}|$. Therefore, LaNAS is more effective than Bayesian methods in high-dimensional tasks.

\begin{table*}[!tb]
  \centering
  \caption{Results on CIFAR-10 using the NASNet search space. LaNet-S and LaNet-L share the same structure, except that the filter size of LaNet-S is 36, while LaNet-L is 128.}
  \begin{tabular}{l l l l l l l}
    \toprule
         \textbf{Model} & \begin{tabular}{@{}c@{}} \textbf{Data} \\  \textbf{Augmentation} \end{tabular} & \begin{tabular}{@{}c@{}} \textbf{Extra} \\  \textbf{Dataset} \end{tabular} & \textbf{Params}  & \textbf{Top1 err}  & \textbf{M} &\textbf{GPU days}\\
    	 \midrule
    	     	 \multicolumn{6}{c}{search based methods} \\
	    \midrule
		  NASNet-A~\cite{zoph2018learning}              & c/o & X  & 3.3 M    & 2.65               & 20000  &  2000         \\
		  AmoebaNet-B-small~\cite{real2019regularized}  & c/o & X  & 2.8  M   & $2.55_{\pm0.05}$   & 27000  &  3150       \\
		  AmoebaNet-B-large                             & c/o & X  & 34.9 M   & $2.13_{\pm0.04}$   & 27000  &  3150 \\
		  PNASNet-5~\cite{liu2018progressive}           & c/o & X  & 3.2 M    & $3.41_{\pm0.09}$   & 1160   &  225         \\
		  NAO~\cite{luo2018neural}                      & c/o & X  & 128.0 M  & 2.11               & 1000   &  200  \\
          EfficientNet-B7                               & c/m+autoaug & ImageNet     & 64M      & 1.01     & ~      & ~     \\
          GPipe~\cite{huang2018gpipe}                   & c/m+autoaug & ImageNet     & 556M     & 1.00     & ~      & ~     \\
          BiT-S                                         & c/m+autoaug & ImageNet     & ~        & 2.49     & ~      & ~     \\
          BiT-M                                         & c/m+autoaug & ImageNet-21k & 928M     & 1.09     & ~      & ~     \\
          \textbf{LaNet-S}     & c/o         & X       & 3.2 M    & {$\mathbf{2.27}_{\pm0.03}$} & 800  & 150   \\
          \textbf{LaNet-S}     & c/m+autoaug & X       & 3.2 M    & {$\mathbf{1.63}_{\pm0.05}$} & 800  & 150   \\
          \textbf{LaNet-L}     & c/o         & X       & 44.1 M   & {$\mathbf{1.53}_{\pm0.03}$} & 800  & 150   \\
		  \textbf{LaNet-L}     & c/m+autoaug & X       & 44.1 M   & {$\mathbf{0.99}_{\pm0.02}$} & 800  & 150   \\
	\midrule
    	 \multicolumn{6}{c}{one-shot NAS based methods} \\
	\midrule
		  ENAS~\cite{pham2018efficient}             & c/o         & X   & 4.6 M    & 2.89               &  -     &  0.45 \\
  		  DARTS~\cite{liu2018darts}                 & c/o         & X   & 3.3 M    & $2.76_{\pm0.09}$   & -      &  1.5  \\
  		  BayesNAS~\cite{zhou2019bayesnas}          & c/o         & X   & 3.4 M    & $2.81_{\pm0.04}$   & -      &  0.2  \\
  		  P-DARTS~(\cite{chen2019progressive})      & c/o         & X   & 3.4 M    & 2.5                & ~ & 0.3 \\
  		  PC-DARTS~(\cite{xu2019pc})                & c/o         & X   & 3.6 M    & $2.57_{\pm0.07}$   & ~ & 0.3 \\
  		  CNAS~(\cite{guo2020breaking})             & c/o         & X   & 3.7 M    & $2.6_{\pm0.06}$    & ~
  		    & 0.3 \\
  		  FairDARTS~(\cite{chu2020fair})            & c/o         & X   & 3.32 M    & $2.54_{\pm0.05}$  & ~
  		    & 3 \\
  		  ASNG-NAS~\cite{akimoto2019adaptive}       & c/o+autoaug & X   & 3.9 M    & $2.83_{\pm0.14}$             & - & 0.11 \\
  		  XNAS+c/0~\cite{nayman2019xnas}            & c/o+autoaug & X   & 3.7 M    & 1.81                         & ~ & 0.3 \\
  		 \textbf{oneshot-LaNet-S}                   & c/o         & X   & 3.6 M    & {$\mathbf{2.24}_{\pm0.02}$}  & - & 3 \\
  		 \textbf{oneshot-LaNet-S}                   & c/o+autoaug & X   & 3.6 M    & {$\mathbf{1.68}_{\pm0.06}$}  & - & 3 \\
  		 \textbf{oneshot-LaNet-L}                   & c/o         & X   & 45.3 M   & {$\mathbf{1.88}_{\pm0.04}$}  & - & 3 \\
  		 \textbf{oneshot-LaNet-L}                   & c/o+autoaug & X   & 45.3 M   & {$\mathbf{1.20}_{\pm0.03}$}   & - & 3 \\
    \bottomrule
    M: number of samples selected. \\
    c/m: cutmix, c/o: cutout \\
    autoaug: auto-augmentation \\
  \end{tabular} 
  \label{open-domain-cifar}
\end{table*}

\begin{table}[!tb]
  \setlength{\tabcolsep}{0.2em}
  \centering
  \caption{Transferring LaNet from CIFAR-10 to ImageNet using the NASNet search space.}
    \begin{tabular}{ l l l l l l l}
    \toprule
          \textbf{Model} &  \textbf{FLOPs}      & \textbf{Params}    & \textbf{top1 err} \\
    	  \midrule  
		  NASNet-C      ~(\cite{zoph2018learning})     & 558M & 4.9 M  & 27.5  \\
		  AmoebaNet-C   ~(\cite{real2018regularized})  & 570M & 6.4 M  & 24.3 \\
		  RandWire      ~(\cite{xie2019exploring})     & 583M & 5.6 M  & 25.3 \\ 
		  PNASNet-5     ~(\cite{liu2018progressive})   & 588M & 5.1 M  & 25.8 \\
		  DARTS         ~(\cite{liu2018darts})         & 574M & 4.7 M  & 26.7 \\
		  BayesNAS      ~(\cite{zhou2019bayesnas})     & -    & 3.9 M  & 26.5 \\
		  P-DARTS       ~(\cite{chen2019progressive})  & 557M & 4.9 M  & 24.4 \\
		  PC-DARTS      ~(\cite{xu2019pc})             & 597M & 5.3 M  & 24.2 \\
		  CNAS          ~(\cite{guo2020breaking})      & 576M & 5.3 M  & 24.6 \\
	\midrule
		  \textbf{LaNet}          & 570M & 5.1 M  & \textbf{23.5}   \\
		  \textbf{oneshot-LaNet}  & 567M & 5.4 M  & \textbf{24.1}   \\
    \bottomrule
    \label{exp:imagenet-transfer-results}
  \end{tabular}
    
\end{table}

\begin{table}[!tb]
  \setlength{\tabcolsep}{0.2em}
  \centering
\caption{Results on ImageNet using the EfficientNet search space. The search cost of LaNet is lower than the cost in Table.1 for directly reusing the supernet from OFA.The LaNet architecture can be found at Table.~\ref{efficientNet-LaNet-Archs} in Appendix.}
  \begin{tabular}{ l l l l l l}
    \toprule
          \textbf{Model} 					  & \textbf{FLOPs} & \textbf{Params} & GPU days & \textbf{top1 err} \\
          \midrule
          FairDARTS     ~(\cite{chu2020fair}) & 440M & 4.3 M  & 3 & 24.4 \\
          FBNetV2-C     ~(\cite{FBNet})       &    375M & 5.5 M & 8.3 & 25.1  \\
          MobileNet-V3  ~(\cite{howard2019searching}) & 219M & 5.8 M &  & 24.8  \\  
          OFA~(\cite{cai2019once})            &    230M & 5.4 M & 1.6 & 23.1 \\
          FBNetV2-F4(~\cite{wan2020fbnetv2})  &    238M & 5.6 M & 8.3 & 24.0 \\
          FairDARTS~(\cite{chu2020fair})      &    386M & 5.3 M & 3 & 22.8 \\
          BigNAS~(\cite{yu2020bignas})        &    242M & 4.5 M & & 23.5 \\
          \midrule
          \textbf{LaNet}                      &    228M & 5.1 M  & 0.3 & \textbf{22.3}   \\
          \midrule
		  OFA~(\cite{cai2019once})           &    595M & 9.1 M & 1.6 & 20.0 \\
		  BigNAS~(\cite{yu2020bignas})        &   586M & 6.4 M & & 23.5 \\
		  FBNetV3~(\cite{dai2020fbnetv3})    &    544M &       & & 20.5 \\
	\midrule
		  \textbf{LaNet}                     &    598M & 8.2 M & 0.3 & \textbf{19.2}   \\
    \bottomrule
    \label{exp:imagenet-efficient-results}
  \end{tabular}
\end{table}

\subsection{Using LaNAS in practice}
\label{lanas_in_practice}
To be consistent with existing literature, we also evaluate \abbr~in searching for architectures for CIFAR-10 using the NASNet search space, and in searching for architectures for ImageNet using the EfficientNet search space.

In sec~\ref{exp:evaluations}, LaNAS allows for repeated samples to be consistent with baselines, which is not desired in practice. Better performance are observable after removing existing samples $D_t$ from search space $\Omega$ in optimizing the acquisition, i.e. max$_{\mathbf{a}_i}$ $\phi(\mathbf{a}_i), \forall\mathbf{a}_i\in \Omega \backslash D_t$, and LaNAS uses this logic through the rest of paper. Therefore, in the following ablation studies (sec.~\ref{exp:ablations}), LaNAS can find $v^{*}$ on NASBench much faster than the case in Fig.~\ref{fig:perf-eval-datasets}.

On CIFAR-10, our search space is same as NASNet~(\cite{zoph2018learning}). We used operations of 3x3 max pool, 3x3, 5x5, depth-separable conv, and skip connection. The search target is the architectures for a reduction and a normal cell, and the number of nodes within a cell is 5. This formulates a search space of $3.5\times10^{21}$ architectures. On ImageNet, our search space is consistent with Efficient-Net. The depth of an Inverted Residual Block (IRB) can be 2, 3, 4; and the expansion ratio within an IRB can be {3, 5, 7}. Therefore, the total possible architectures are around $10^{20}$.

The setup of supernet on CIFAR-10 is consistent with the description in sec.\ref{oneshot-nas}; on ImageNet, we reused supernet from~\cite{cai2019once}. We selected the top architecture collected from the search and re-trained them for 600 epochs to acquire the final accuracy in Table.~\ref{open-domain-cifar}. We reused the training logic from \texttt{DARTS} and their training settings.

Table.~\ref{open-domain-cifar} compares our results in the context of searching NASNet style architecture on CIFAR-10. The best performing architecture found by LaNAS in 800 samples demonstrates an average accuracy of 98.37\% (\#filters = 32, \#params = 3.22M) and 99.01\% (\#filters = 128, \#params = 44.1M), which is better than other results on CIFAR-10 without using ImageNet or transferring weights from a network pre-trained on ImageNet. It is worth noting that we achieved this accuracy with 33x fewer samples than AmoebaNet. Besides, one-shot LaNAS also consistently demonstrates the strongest result among other one-shot variants in similar GPU time. Besides, the results on ImageNet also consistently outperform SoTA models.

The performance gap between one-shot NAS and search based NAS is largely due to inaccurate predictions of $v_i$ from supernet~\cite{sciuto2019evaluating, zhao2020few}. Improving supernet is beyond this work, and we provide a solution at~\cite{zhao2020few}.

\begin{figure}[t]
  \begin{center}
    \includegraphics[height=4.0cm]{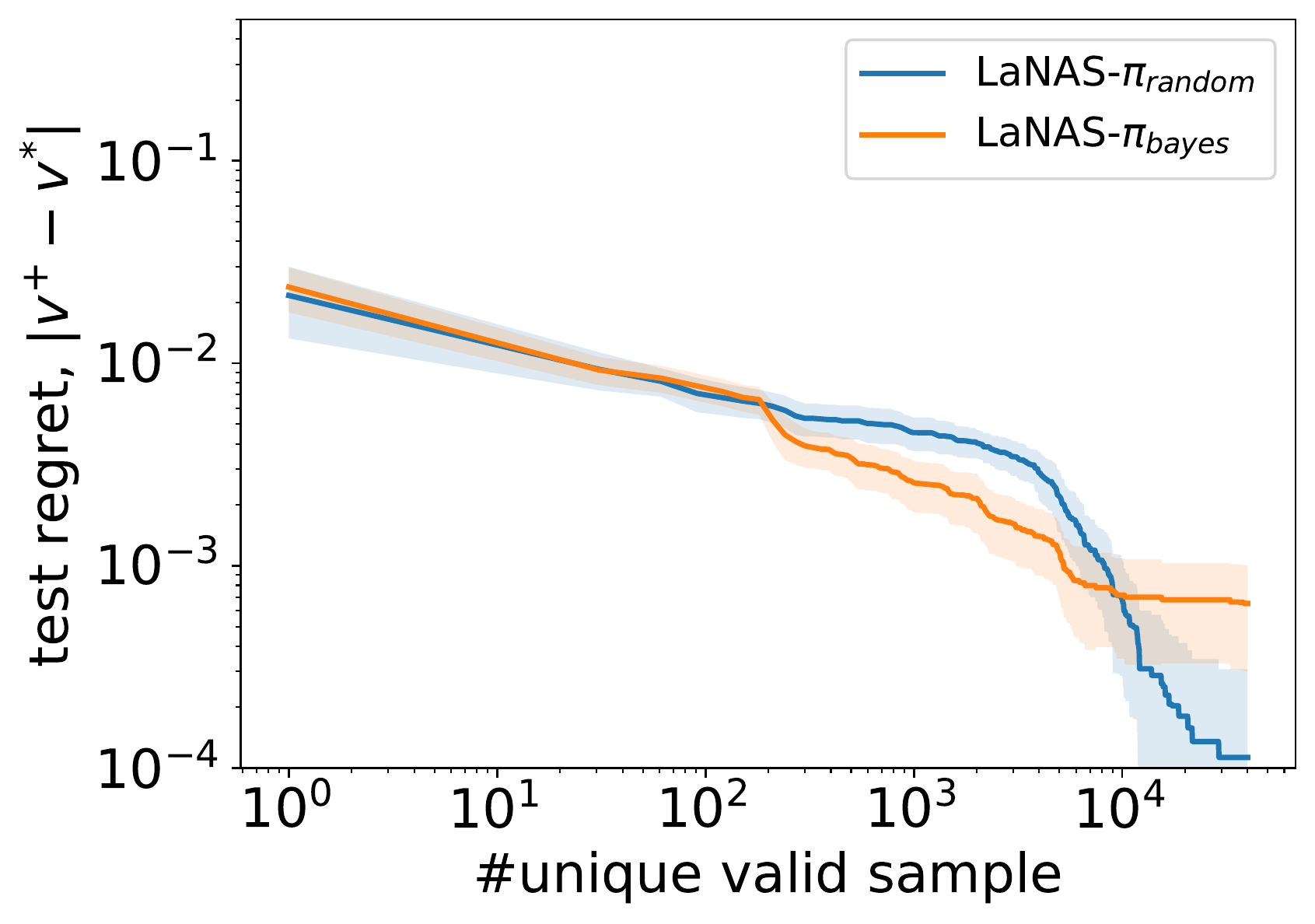}
  \end{center}
    \caption{Comparisons of $\pi_{bayes}$ to $\pi_{random}$ in sampling from the selected partition $\Omega_j$ }
    \label{fig:sampleing-policy}
\end{figure}
\label{exp:ablations}

\begin{figure*}[h]
\begin{center}
\includegraphics[width=0.9\textwidth]{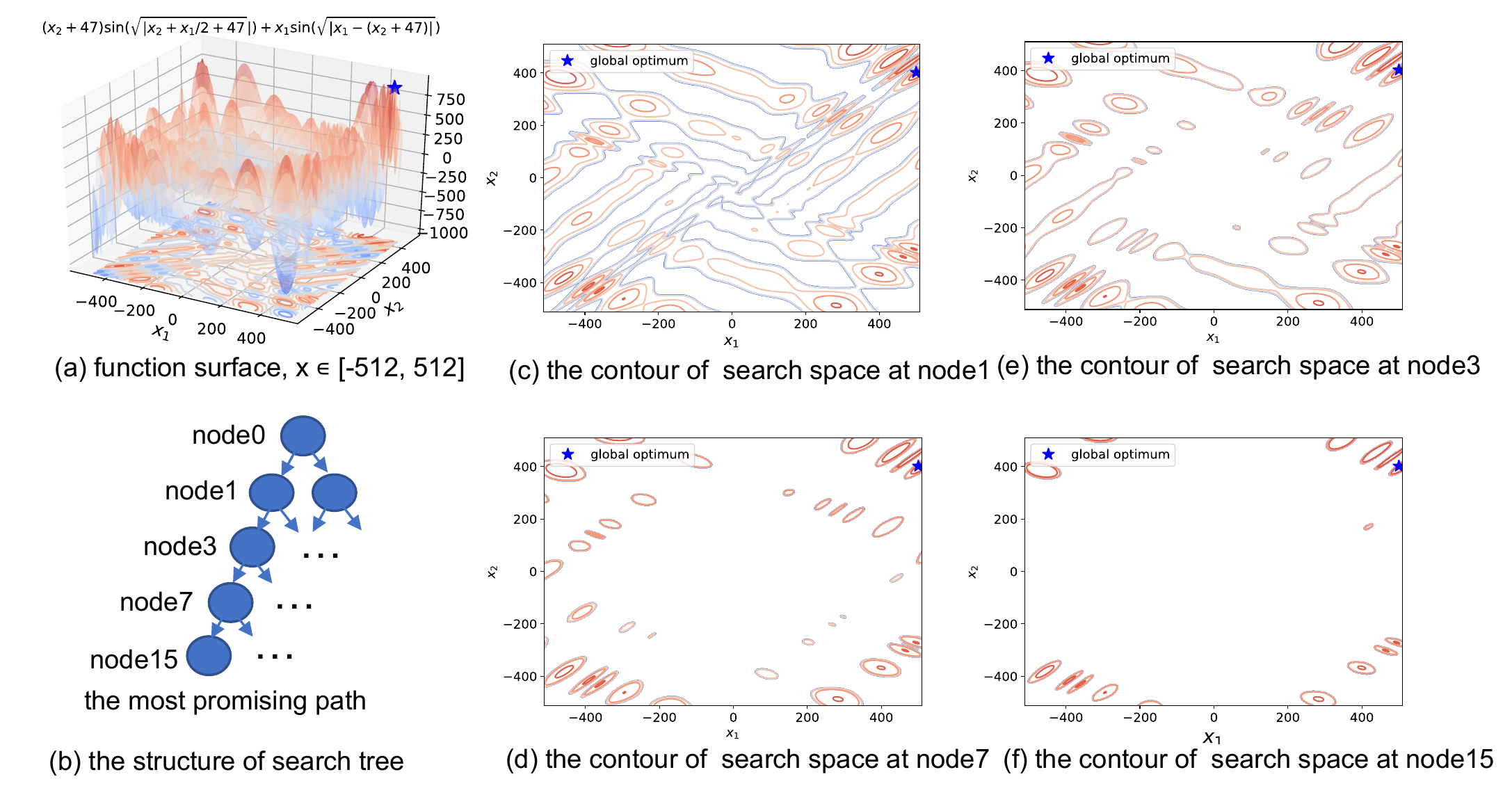}
\end{center}
\caption{\textbf{A visualization of partitioning eggholder function using LaNAS}: eggholder is a popular benchmark function for black-box optimization, (a) depicts its function surface, contour and definition. (b) LaNAS builds a tree of height = 5 for searching $v^{*}$; after collecting 500 samples, we visualize each partitions $\Omega_j$ represented at node0$\rightarrow$node15 in (c)$\rightarrow$(f), by splitting $\Omega$ based on its parent constraint. As node0$\rightarrow$node15 recursively splitting $\Omega$, the final $\Omega_j$ at node15 only contains the most promising region in $\Omega$ for sampling (see (f) v.s. (c)), and the bad region (blue lines in contours) are clearly separated from good region (red lines in contours) from (c)$\rightarrow$(f).}
\label{partition-viz}
\end{figure*}

\subsubsection{Hyper-parameters tuning in LaNAS}

\textbf{The effect of tree height and \#selects}: Fig.~\ref{fig:ablation_study}(a) relates tree height ($h$) and the number of selects (\#selects) to the search performance. Each entry represents \#samples to find $v^{*}$ on NASBench, averaged over 100 runs. A deeper tree leads to better performance since the model space is partitioned by more leaves. Similarly, small \#select results in more frequent updates of action space allowing the tree to make up-to-date decisions, and thus leads to improvement. On the other hand, the number of classifiers increases exponentially as the tree goes deeper, and a small \#selects incurs a frequent learning phase. Therefore, both can significantly increase the computation cost.

\textbf{Choice of classifiers}: Fig.\ref{fig:ablation_study}(b) shows that using a linear classifier performs better than an multi-layer perceptron (MLP). This indicates that adding complexity to the decision boundary of actions may not help with the performance. Conversely, performance can get degraded due to potentially higher difficulties in optimization.

\textbf{\#samples for initialization}: We need to initialize each node classifier properly with a few samples to establish the initial boundaries. Fig.\ref{fig:ablation_study}(c) shows cold start is necessary (init $>$ 0 is better than init = 0), and a small init=100-400 converges to top 5\% performance much faster than init=2000, while init=2000 gets the best performance a little faster.

\textbf{The effect of $c$ in UCB}: Fig.~\ref{fig:ablation_study}(d) shows that the exploration term, $c\sqrt{\frac{log(n_{curt})}{n_{next} }}$, improves the performance as $c$ increases from 0 to 0.1, while using a large $c$, e.g. $>$ 0.5, is not desired for over-exploring. Please noted the optimal $c$ is small as the maximum accuracy = 1. In practice, we find that setting $c$ to $0.1*$max accuracy empirically works well. For example, if a performance metric in the range of [0, 100], we recommend setting c = 10. 

\textbf{Using $\pi_{bayes}$ v.s. $\pi_{random}$ for sample proposal}: though $\pi_{bayes}$ is faster in the beginning, $\pi_{random}$ delivers the better final result due to the consistent random exploration in the most promising partition. Therefore, we used $\pi_{random}$ for simplicity and good final performance through this paper.

\subsection{Analysis of LaNAS}
\label{analysis_search}

\textbf{Experiment design}: To validate the effectiveness of latent actions in partitioning the search space into regions with different performance metrics, and to visualize the search phase of LaNAS, we look into the dynamics of sample distributions on tree leaves during the search. By construction, left nodes contain regions of the good metric while the right nodes contain regions of the poor metric. Therefore, at each node $j$, we can construct \emph{reference} distribution $p^*_j(v)$ by training toward a NAS dataset to partition the dataset into small regions with concentrated performances on leaves, i.e. using regression tree for the classification. Then, we compare $p^*_j(v)$ with the estimated distribution $p^n_j(v)$, where $n$ is the number of accumulated samples in $D_t\cap\Omega_j$ at the node $j$ at the search step $t$. Since the \emph{reference} distribution $p^*_j(v)$ is static, visualizing $p^n_j(v)$ to $p^*_j(v)$ and calculating $D_{KL}[p^n_j \|\| p^*_j]$ enables us to see variations of the distribution over partition $\Omega_j$ on tree leaves w.r.t growing samples to validate the effectiveness of latent actions and to visualize the search.

\textbf{Experiment setup}: we used NASBench-101 that provides us with the true distribution of model accuracy, given any subset of model specifications, or equivalently a collection of actions (or constraints). In our experiments, we use a complete \emph{binary} tree with the height of $5$. We label nodes 0-14 as internal nodes, and nodes 15-29 as leaves. By definition, $\bar v^*_{15} > \bar v^*_{16} ... > \bar v^*_{29}$ reflected by $p^*_{15,16,28,29}$ in Fig.~\ref{fig:search-dynamics}b.

\textbf{Explanation to the performance of LaNAS}: at the beginning of the search ($n = 200$ for random initialization), $p^{200}_{15,16}$ are expected to be smaller than $p^*_{15,16}$, and $p^{200}_{28, 29}$ are expected to be larger than $p^*_{15,16}$; because the tree still learns to partition at that time. With more samples ($n=700$), $p_j$ starts to approximate $p^*_j$, manifested by the increasing similarity between $p^{700}_{15,16,28,29}$ and $p^*_{15,16,28,29}$, and the decreasing $D_{KL}$ in Fig.~\ref{fig:search-dynamics}a. This is because MCTS explores the under-explored regions, and it explains the comparable performance of LaNAS to baselines in Fig.~\ref{fig:perf-eval-datasets}. As the search continues ($n\rightarrow 5000$), {\abbr} explores deeper into promising regions and $p^n_j$ is biased toward the region with good performance, deviated from $p^*_j$. As a result, $D_{KL}$ bounces back in Fig.~\ref{fig:search-dynamics}a. These search dynamics show how our model adapts to different stages during the course of the search, and validate its effectiveness in partitioning the search space.

\textbf{Effective partitioning}:  The mean accuracy of $p_{15}^{700, 5000} > p_{16}^{700, 5000} > p_{28}^{700, 5000} > p_{29}^{700, 5000}$ in Fig.~\ref{fig:search-dynamics}(b) indicates that {\abbr} successfully minimizes the variance of rewards on a search path making architectures with similar metrics concentrated in a region, and {\abbr} correctly ranks the regions on tree leaves. These manifest that LaNAS fulfills the online partitioning of $\Omega$. An example partitioning of 2d egg-holder function can be found in Fig.~\ref{partition-viz}.

\section{Related works}
\label{related-works}

Sequential Model Based Optimizations (SMBO) is a classic black box optimization framework~\cite{hutter2009automated, hutter2011sequential}, that uses a surrogate $S$ to extrapolate unseen region in $\Omega$ and to interpolate the explored region with existing samples. In the scenarios of expensive function evaluations $f(\mathbf{a}_i)$, SMBO is quite efficient by approximating $f(\mathbf{a}_i)$ with $S(\mathbf{a}_i)$. SMBO proposes new samples by solving max$_{\mathbf{a}_i}\phi(\mathbf{a}_i)$ on $S$, where $\phi$ is a criterion, e.g. Expected Improvement (EI)~\cite{jones2001taxonomy} or Conditional Entropy of the Minimizer (CEM)~\cite{villemonteix2009informational}, that transforms the value predicted from $S$ for better trade-off between exploration and exploitation.

Bayesian Optimization (BO) is also an instantiation of SMBO~\cite{ wang2013bayesian, gardner2014bayesian} that utilizes a Gaussian Process Regressor (GPR) as $S$. However, GPR suffers from the issue of $\mathcal{O}(n^3)$ where $n$ is \#samples. To resolve these issues, \cite{hutter2011sequential} replaces GPR with random forests, called SMAC-random forest, to estimate $\hat{\mu}$ and $\hat{\sigma}$ for predictive Gaussian distributions, and \cite{bergstra2011algorithms} proposes Tree-structured Parzen Estimator (TPE) in modeling Bayesian rule. Though both resolves the cubic scaling issue, as we thoroughly explained in sec~\ref{exp:alg-reasoning}, the key limitation of Bayesian approaches is at auxiliary optimization of acquisition function on an intractable search space. Similarly, recent predictor-based methods~\cite{shi2019multi} have achieved impressive results on NASBench by predicting every unseen architecture from the dataset. Without predicting over the entire dataset, their performance can drastically deteriorate. LaNAS eliminates this undesired constraint in BO or predictor-based methods, being scalable regardless of problem dimensions, while still captures promising regions for sample proposals.

Besides the recent success in games~\cite{silver2016mastering, tian2019elf}, Monte Carlo Tree Search (MCTS) has also been used in robotics planning, optimization, and NAS~\cite{bucsoniu2013optimistic, munos2014bandits, weinstein2012bandit, mansley2011sample}. AlphaX is the first MCTS based NAS agent to explore the search space assisted with a value function predictor. However, the action space of AlphaX is manually defined w.r.t the search space. In sec.~\ref{sec:motivation}, we have clearly demonstrated that manually defined search space provides a confusing reward signal to the search, therefore low sample efficiency. In contrast, LaNAS learns the action space to partition the search space into good and bad regions, providing a stronger reward signal to guide the search; and \cite{wang2020learning} extends LaNAS to be a generic black box meta-solver.

XNAS~\cite{nayman2019xnas} and many other existing NAS methods in Table.~\ref{open-domain-cifar} are specifically designed to improve the search by exploiting the architecture characteristics in the NASNet or EfficientNet search space, while LaNAS is a new generic search algorithm applicable to the much broader scope of tasks. For example, P-DARTS~\cite{chen2019progressive} observes there is a depth gap in the architecture during the search and evaluation steps in the search for CNN on the NASNet search space. Then, P-DARTS proposes progressively increasing the network depth during the search for CNN, so that the network depth can match the evaluation setting. But these task-dependent settings can change from tasks to tasks, or search space to search space. LaNAS treats NAS as a black box function without making any assumptions to the underlying search space, being adaptable to different problems. In the black box optimization challenge at NeurIPS-2020, LaNAS is proven to be effective in solving 216 different ML tasks~\cite{sazanovich2020solving}.

\section{Conclusion}
This work presents a novel MCTS based search algorithm that learns action space for MCTS. With its application to NAS, LaNAS has proven to be more sample-efficient than existing approaches, validated by the cases with and without one-shot NAS on a diverse of tasks. The proposed algorithm is not limited to NAS and has been extended to be a generic gradient-free algorithm~\cite{wang2020learning}, applied to different challenging black-box optimizations.

\bibliography{lanas_icml}
\bibliographystyle{IEEEtran}

\newpage

\begin{IEEEbiography}[{\includegraphics[width=1in,height=1in,clip,keepaspectratio]{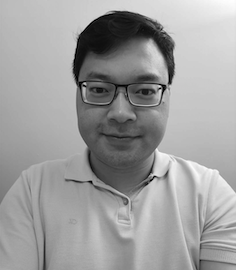}}]{Linnan Wang}
Linnan is a Ph.D. student at the CS department of Brown University, advised by Prof.Rodrigo Fonseca. Before Brown, he was a OMSCS student at Gatech while being a full time software developer at Dow Jones. He acquired his bachelor degree from University of Electronic Science and Technology of China (UESTC). His research interests include Artificial Intelligence (AI) and High Performance Computing (HPC). 
\end{IEEEbiography}

\begin{IEEEbiography}[{\includegraphics[width=1in,height=1.25in,clip,keepaspectratio]{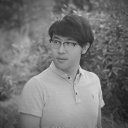}}]{Saining Xie}
Saining is a research scientist at Facebook AI Research. He was a PhD student at CSE Department, UC San Diego, advised by Professor Zhuowen Tu. He obtained his bachelor degree from Shanghai Jiao Tong University in 2013. His research interest includes machine learning (especially deep learning) and its applications in computer vision.
\end{IEEEbiography}

\begin{IEEEbiography}[{\includegraphics[width=1in,height=1.25in,clip,keepaspectratio]{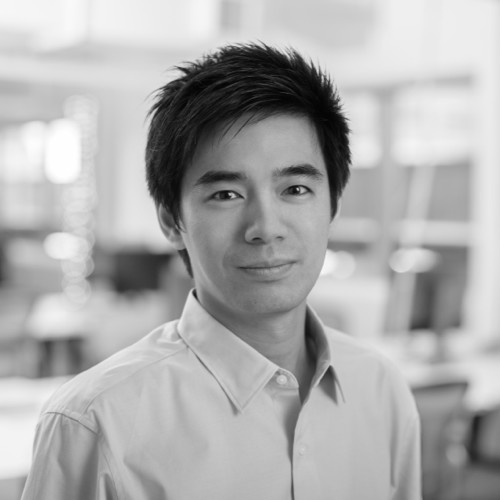}}]{Teng Li}
Teng is a a research scientist at Facebook AI Research. He received my PhD in computer engineering from the George Washington University in Washington, DC. His PhD research is primarily within the area of GPGPU, parallel and distributed systems, and high-performance computing.
\end{IEEEbiography}

\begin{IEEEbiography}[{\includegraphics[width=1in,height=1.25in,clip,keepaspectratio]{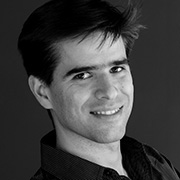}}]{Rodrigo Fonseca}
Rodrigo is an associate professor at Brown University's Computer Science Department. His work revolves around distributed systems, networking, and operating systems. Broadly, He is interested in understanding the behavior of systems with many components for enabling new functionality, and making sure they work as they should.
\end{IEEEbiography}

\begin{IEEEbiography}[{\includegraphics[width=1in,height=1.25in,clip,keepaspectratio]{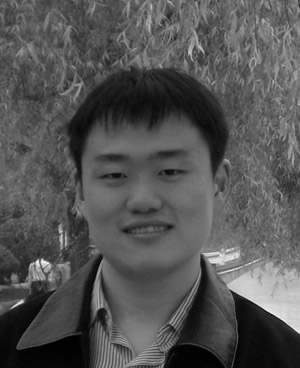}}]{Yuandong Tian}
 Yuandong is a Research Scientist in Facebook AI Research (FAIR). He received Ph.D in the Robotics Institute, Carnegie Mellon University, advised by Srinivasa Narasimhan. He received Master and Bachelor degrees in Computer Science and Engineering Department, Shanghai Jiao Tong University. He is a recipient of ICCV 2013 Marr Prize Honorable Mentions for a hierarchical framework that gives globally optimal guarantees for non-convex non-rigid image deformation.
\end{IEEEbiography}

\clearpage

\section{Appendix}
\subsection{LaNAS algorithms}

\begin{algorithm}[h]
   \caption{LaNAS search procedures }
   \label{alg:lanas_procedures}
  \begin{algorithmic}[1]
    \WHILE{ acc $<$ target } 
         \FOR{ $n\in Tree.N$ } 
            \STATE { $n.g.train()$ } 
         \ENDFOR
         
         \FOR{$i = 1 \rightarrow \#selects$}
            \STATE $leaf, path = ucb\_select(root) $
            \STATE $constraints = get\_constraints(path)$
            \STATE $network = sampling(constraints)$
            \STATE $acc = network.train()$
            \STATE $back\_propagate(network, acc)$
         \ENDFOR
    \ENDWHILE
  \end{algorithmic}
\end{algorithm}

\begin{algorithm}[h]
   \caption{get\_constraints(path) in Alg.~\ref{alg:lanas_procedures} }
   \label{alg:get_consts}
  \begin{algorithmic}[1]
  
    \STATE	$constraints$ = [] \\
		\FOR{ $node \in s\_path$ } 
			\STATE $\mathbf{W}, b = node.g.params()$
			\STATE $\bar{X} = node.\bar{X}$
			\IF{ $node$ on left } 
                \STATE $constraints.add( \mathbf{W}\mathbf{a} +b \geq \bar{X} )$
            \ELSE 
                \STATE $constraints.add(\mathbf{W}\mathbf{a} +b < \bar{X} )$
            \ENDIF
		\ENDFOR
		\RETURN $constraints$
  \end{algorithmic}
\end{algorithm}

\begin{algorithm}[h]
   \caption{ucb\_select($c = root$) in Alg.~\ref{alg:lanas_procedures} }
   \label{alg:ucb_select}
  \begin{algorithmic}[1]
    \STATE $path = []$
        \WHILE{$c$ not $leaf$} 
            \STATE{$path.add(c)$}
            \STATE{$l_{ucb}= get\_ucb(c.left.\bar{X}, c.left.n, c.n)$}
            \STATE{$r_{ucb}= get\_ucb(c.right.\bar{X}, c.right.n, c.n)$}
        \ENDWHILE

		\WHILE{ $c$ not $leaf$ }
			 \STATE $path.add(c)$ \\
			 \IF{ $l_{ucb} > r_{ucb}$ } 
			    \STATE {$c = c.left$} 
			 \ELSE 
			    \STATE{$c = c.right$} 
			 \ENDIF
	    \ENDWHILE
		\RETURN $path, c$
  \end{algorithmic}
\end{algorithm}

\begin{algorithm}[h]
\caption{get\_ucb($\bar{X}_{next}$, $n_{next}$, $n_{curt}$ ) in Alg.~\ref{alg:ucb_select} }
   \label{alg:ucb}
   \begin{algorithmic}[1]
    \STATE c = 0.1
    \IF{ $n_{next} = 0$ } 
        \STATE return $+\infty$
    \ELSE 
        \STATE return $\frac{\bar{X}_{next}}{n_{next}}+2c\sqrt{\frac{2log(n_{curt})}{n_{next} }}$
    \ENDIF
  \end{algorithmic}
\end{algorithm}

\begin{figure}[h]
    \centering
     \subfloat[][max$\phi(\mathbf{a}_i)$ using TPE and RE]{ \includegraphics[height=4.0cm]{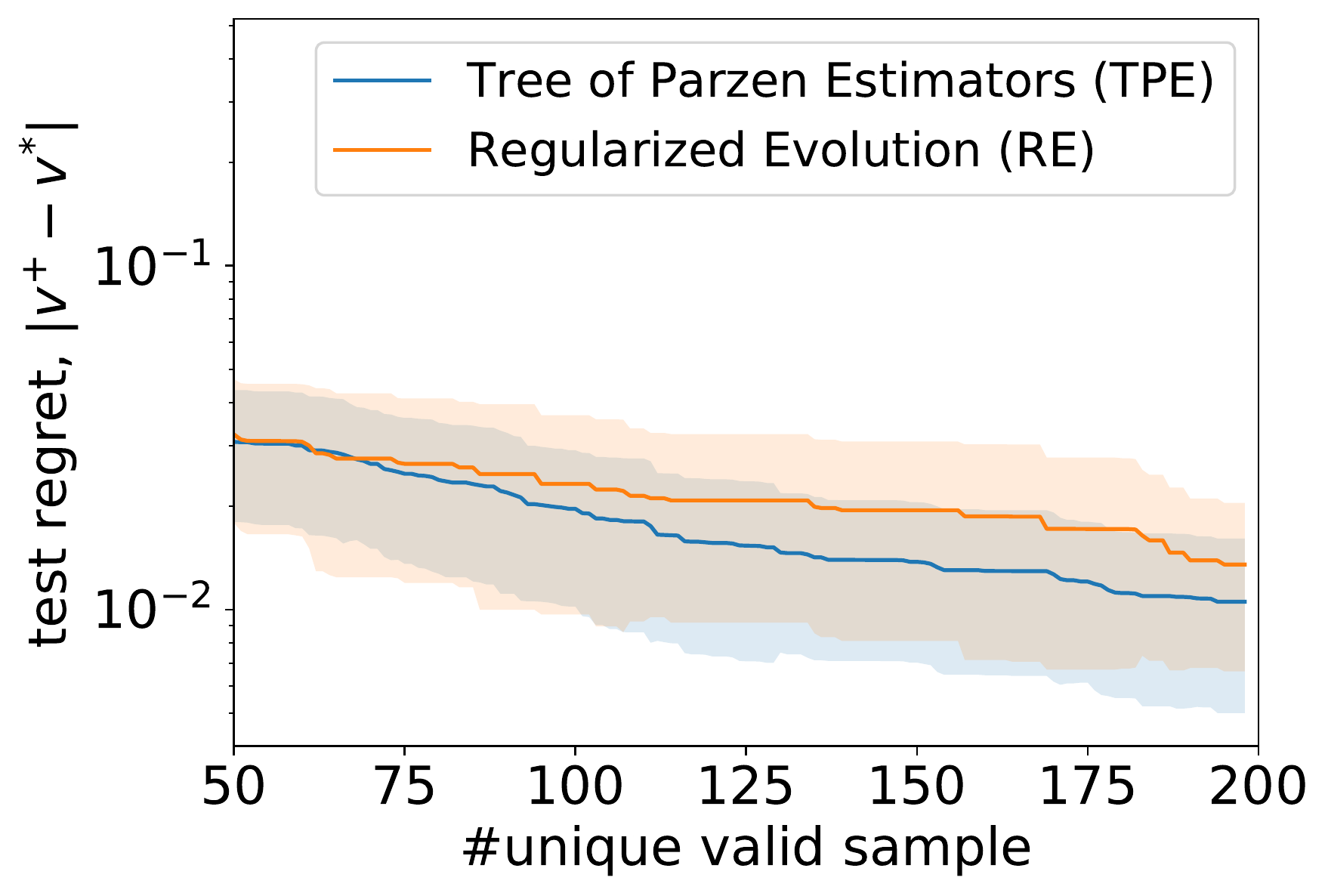} } \\ 
     \subfloat[][max$\phi(\mathbf{a}_i)$ with the hill climbing using 1\%, 10\% and 100\% of search space $\Omega$]{ \includegraphics[height=4.0cm]{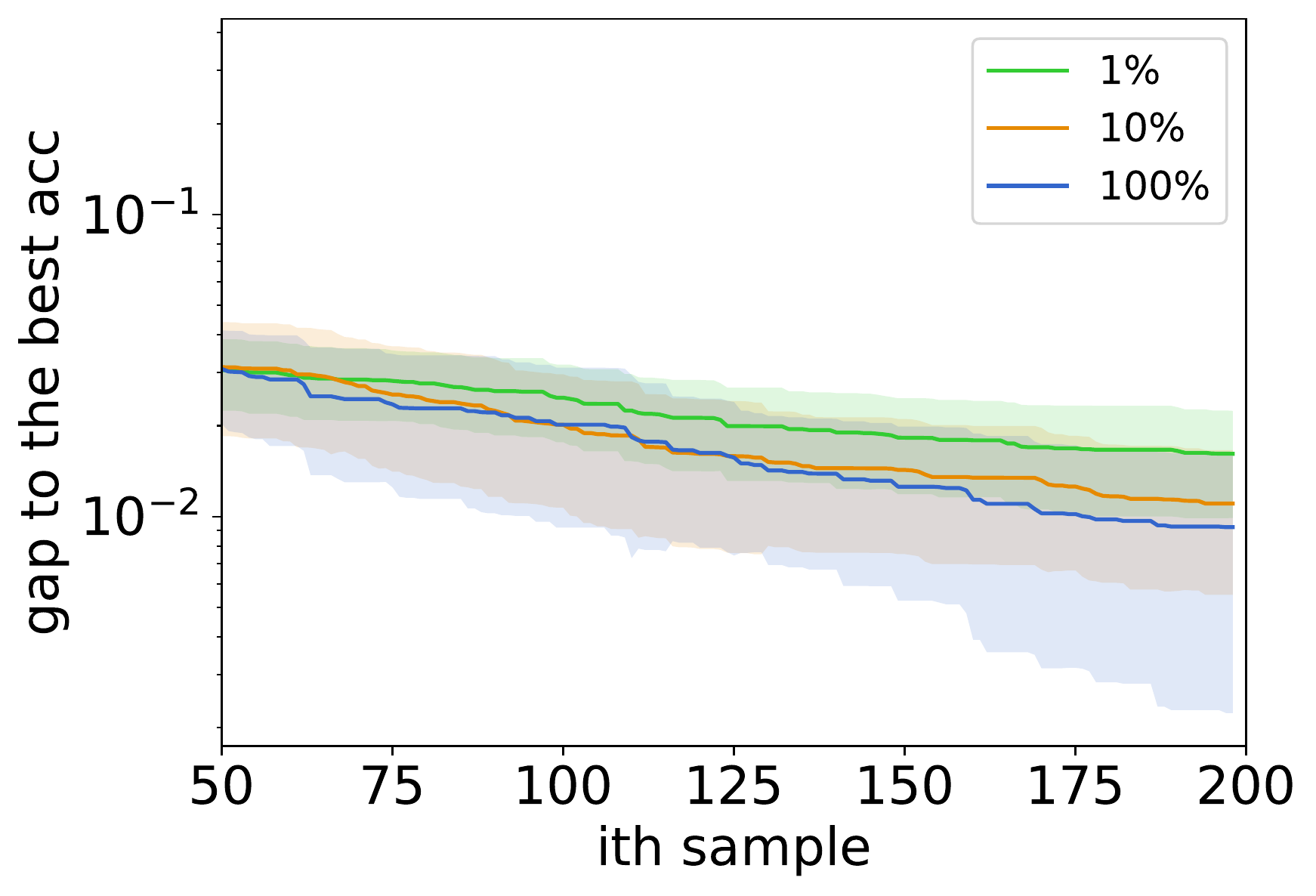} }
 	 \caption{\textbf{The impact of max$\phi(\mathbf{a}_i)$ in the Bayesian Optimization (BO)}: BO proposes samples by a non-convex optimization of the acquisition function, i.e. max$\phi(\mathbf{a}_i)$. \textbf{(a) shows} that the BO performance is closely related to the performance of search methods used in max$_{\mathbf{a}_i}$ $\phi(\mathbf{a}_i)$, and \textbf{(b) shows} the BO performance deteriorates after reducing the search budget for a local search algorithm (hill climbing) from probing $100\%$ search space $\Omega$ to, probing $1\%$ of $\Omega$ in max$\phi(\mathbf{a}_i)$.
 	 In a high dimensional search space, e.g. NAS $|\Omega|\sim 10^{20}$, it is impossible to find the global optimum in max$\phi(\mathbf{a}_i)$, thereby deteriorating the performance of BO methods.
 	 }\label{suboptimal_opt}
\end{figure}

\begin{figure}[h]
    \centering
     \subfloat[][normal cell]{ \includegraphics[width=0.98\columnwidth]{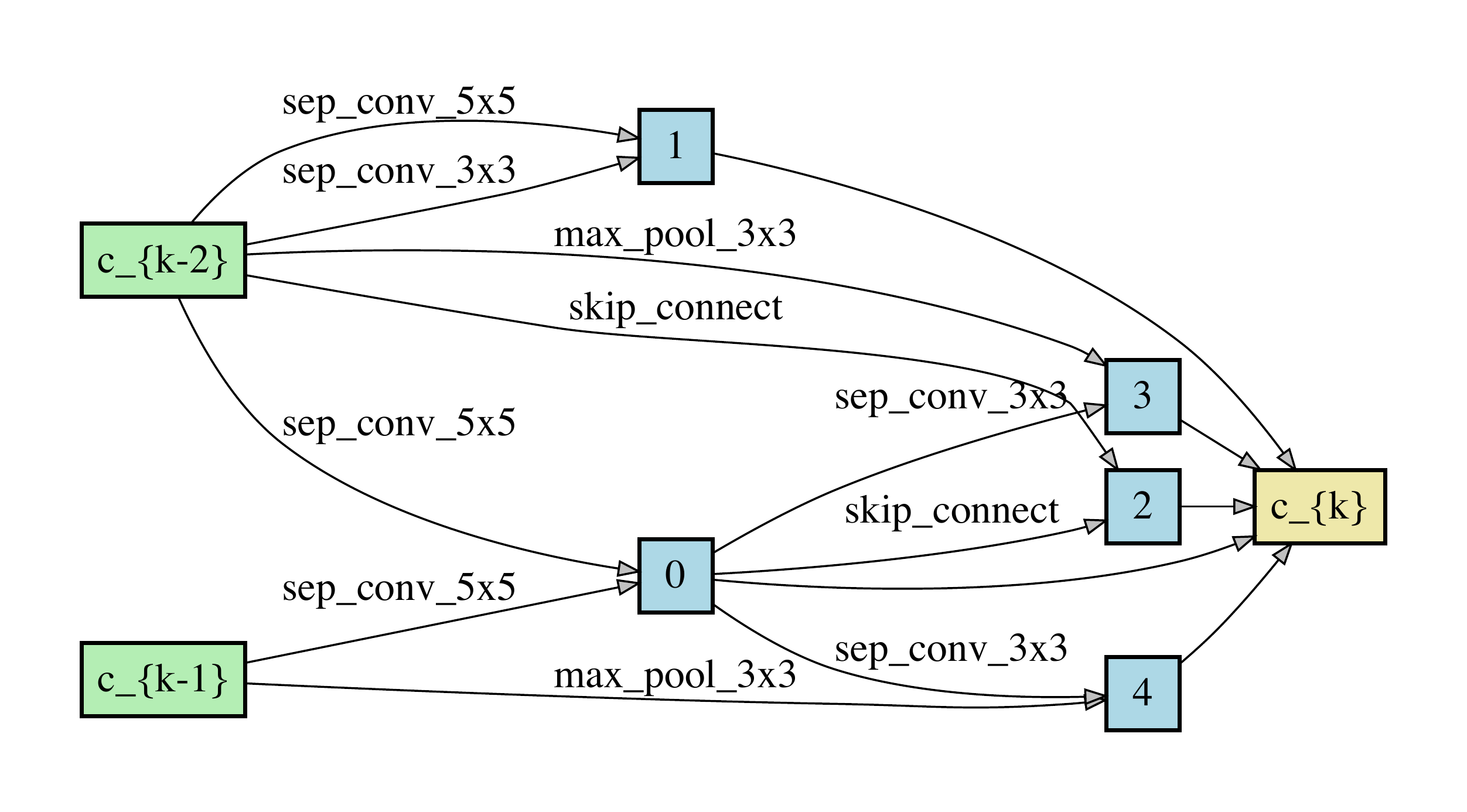} } \\
 	 \subfloat[][reduction cell]{ \includegraphics[width=0.98\columnwidth]{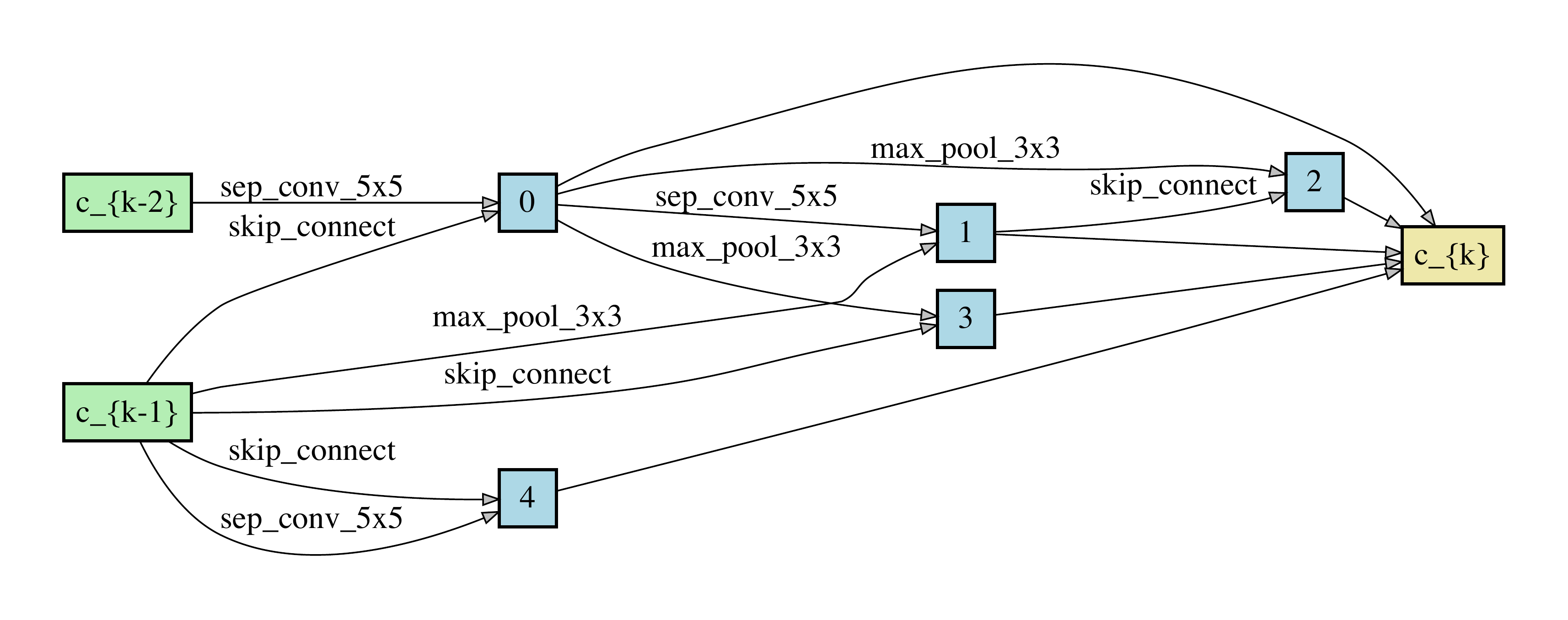} }
 	 \caption{\textbf{searched cell structure}: the best cell structure found in the search.
 	 }\label{cell_structure}
\end{figure}

The best convolutional cell found by LaNAS is visualized in Fig.~\ref{cell_structure}.
For details of constructing a network with learned cells, please refer to Fig.15 in~\cite{wang2019alphax}.

\clearpage

\begin{table}[!tb]
  \centering
  \caption{LaNet architecture found on the EfficientNet search space, i.e. results in Table.~\ref{exp:imagenet-efficient-results}.}
  \begin{tabular}{l l l l l l}
    \toprule
         \textbf{id} & \textbf{block} & \textbf{kernel} & \textbf{stride} & \begin{tabular}{@{}c@{}} \textbf{Output} \\  \textbf{Channel} \end{tabular} & \begin{tabular}{@{}c@{}} \textbf{Expand} \\  \textbf{Ratio} \end{tabular} \\
    	 \midrule
    	 \multicolumn{6}{c}{LaNet@228M FLOPS} \\
         \midrule
    	 0 & Conv & 3 & 2 & 24 &  \\
    	 1 & IRB & 3 & 1 & 24  & 1 \\
    	 \midrule
    	 2 & IRB & 3 & 2 & 36  & 3 \\
    	 3 & IRB & 3 & 1 & 36  & 3 \\
    	 \midrule
    	 4 & IRB & 3 & 2 & 40  & 3 \\
    	 5 & IRB & 3 & 1 & 40  & 4 \\
    	 6 & IRB & 3 & 1 & 40  & 4 \\
    	 \midrule
    	 7 & IRB & 5 & 2 & 80  & 4 \\
    	 8 & IRB & 5 & 1 & 80  & 4 \\
    	 9 & IRB & 3 & 1 & 80  & 4 \\
    	 10 & IRB & 3 & 1 & 112  & 4 \\
    	 11 & IRB & 3 & 1 & 112  & 4 \\
    	 \midrule
    	 12 & IRB & 5 & 2 & 168  & 6 \\
    	 13 & IRB & 4 & 1 & 168  & 6 \\
    	 14 & IRB & 3 & 1 & 168  & 6 \\
    	 15 & IRB & 3 & 1 & 168  & 6 \\
    	 \midrule
    	 16 & Conv & 1 & 1 & 960 &  \\
    	 17 & Conv & 1 & 1 & 1280 &  \\
    	 18 & FC & 1 & 1 & 1000 & \\
    	 \midrule
    	 \multicolumn{6}{c}{LaNet@598M FLOPS} \\
    	 \midrule
    	 0 & Conv & 3 & 2 & 24 &   \\
    	 0 & IRB  & 3 & 1 & 24 & 1 \\
    	 \midrule
    	 0 & IRB  & 3 & 2 & 36 & 3 \\
    	 0 & IRB  & 5 & 1 & 36 & 4 \\
    	 0 & IRB  & 3 & 1 & 36 & 4 \\
    	 \midrule
    	 0 & IRB  & 5 & 2 & 48 & 4 \\
    	 0 & IRB  & 5 & 1 & 48 & 4 \\
    	 0 & IRB  & 3 & 1 & 48 & 6 \\
    	 0 & IRB  & 3 & 1 & 48 & 6 \\
    	 \midrule
    	 0 & IRB  & 7 & 2 & 96 & 6 \\
    	 0 & IRB  & 5 & 1 & 96 & 6 \\
    	 0 & IRB  & 3 & 1 & 96 & 6 \\
    	 0 & IRB  & 5 & 1 & 96 & 6 \\
    	 0 & IRB  & 3 & 1 & 136 & 6 \\
    	 0 & IRB  & 5 & 1 & 136 & 6 \\
    	 0 & IRB  & 5 & 1 & 136 & 6 \\
    	 0 & IRB  & 3 & 1 & 136 & 6 \\
    	 \midrule
    	 0 & IRB  & 3 & 2 & 200 & 6 \\
    	 0 & IRB  & 5 & 1 & 200 & 6 \\
         0 & IRB  & 5 & 1 & 200 & 6 \\
         0 & IRB  & 3 & 1 & 200 & 6 \\
        \midrule
        16 & Conv & 1 & 1 & 1152 &  \\
        16 & Conv & 1 & 1 & 1536 &  \\
        18 & FC & 1 & 1 & 1000 & \\
		 \bottomrule
  \end{tabular} 
  \label{efficientNet-LaNet-Archs}
\end{table}

\clearpage
\section{Architecture characteristics in the good partitions}
\label{architecture-characteristisc}

LaNAS partitions the search space to find sub-regions where contain good architectures. Architectures falling into the most promising sub-region is supposed to demonstrate some of the good design heuristics, e.g. residual connections, deep networks, and many filters. Here we design a controlled study to show the architecture characteristics in the most promising partition learnt from the search. Please note the focus of this paper is at proposing a new search algorithm for NAS, we leave learning new design heuristics on the complex problems as a future work.

To show LaNAS can learn good heuristics from a search space, we perform the experiment on the ConvNet-60K (Fig.~\ref{fig:perf-eval-datasets}), which contains over 60000 the architecture and accuracy pairs. ConvNet design space contains AlexNet style sequential networks. In the design space, the depth of networks ranges from 1 to 5. We only use the convolution for each layers, but the filter size of a convolution layer is in the set of $\{32, 64, 96\}$, and the kernel size is $\{3\times3, 5\times5, 7\times7\}$. Therefore, this design space renders 66429 ConvNets in total. We encode a network by the following rule:
\begin{center}
\begin{tabular}{ c c c }
 kernel & filters & code \\ \midrule
 $3\times3$ & 32  & 0.1 \\
 $3\times3$ & 64  & 0.2 \\
 $3\times3$ & 96  & 0.3 \\
 $5\times5$ & 32  & 0.4 \\
 $5\times5$ & 64  & 0.5 \\
 $5\times5$ & 96  & 0.6 \\
 $7\times7$ & 32  & 0.7 \\
 $7\times7$ & 64  & 0.8 \\
 $7\times7$ & 96  & 0.9 \\
 \multicolumn{2}{c}{empty layer} & 1.0 \\
\end{tabular}
\end{center}
For any networks with the depth $< 5$, we pad the network coding with 1.0 to ensure the coding length = 5. We use a search tree of height = 4, so the most promising partition is enclosed by the 3 intermediate nodes (excluding the leaf) on the left-most path.

Since ConvNet is a well-studied design space; and some design heuristics such as the deep network works better than shallow networks, more filters improves the accuracy are known. If LaNAS works, it is supposed to find those existing heuristics from the most promising partition. We stop LaNAS after it collects 300 samples, and save the parameters of linear regressors $(\mathbf{w}_i, b_i)$ on the left-most path to test on the dataset. Then we use these regressors to find unseen architectures located on the left-most node, by $\mathbf{w}_i*\mathbf{x}_i+b_i > \hat{V}(\Omega_j)$, where $\hat{V}(\Omega_j)$ is the average accuracy of sampled architecture located on the node. We found 3510 architectures out of 65643 architectures satisfy these 3 regressors, i.e. those 3510 networks are located on the most promising partition ($\Omega_{best}$).
  
Here are some common characteristics observed from those 3510 networks predicted in $\Omega_{best}$, after LaNAS collecting 300 samples:
\begin{itemize}
  \item \textbf{Good networks are deep}: among those networks predicted on $\Omega_{best}$, the number of networks with depth = 1,2,3,4,5 are 0, 0, 35, 41, 3434 (3510 in total), while the number of networks at depth = 1,2,3,4,5 on the entire space are 9, 81, 722, 6484, 58347 (65643 in total). To show $\Omega_{best}$ prefers deeper networks, we calculate the ratio of each depth w.r.t the entire number, e.g. the ratio of depth = 5 networks on $\Omega_{best}$ is 3434/3510, while the ratio of depth = 5 networks on $\Omega$ is 58347/65643. From the table below, we can see $\Omega_{best}$ indeed prefers deeper networks as the ratio of depth = 5 networks (0.98) is much larger than the case on the $\Omega$ (0.89).
\begin{center}
\begin{tabular}{ c c c c c c }
 ratio     & depth = 1 & 2 & 3 & 4 & 5 \\  \midrule
 $\Omega_{best}$ & 0.00 & 0.00 & 0.01 & 0.01 & 0.98 \\
 $\Omega$        & 0.0001  & 0.001 & 0.011 & 0.10 & 0.89 \\
\end{tabular}
\end{center}

  \item \textbf{Good networks have more filters}: the design of ConvNet space suggests that the filter size has the equal probability over $\{32, 64, 96\}$ at each layers. Here we calculate the likelihood of each filter sizes for networks located in $\Omega_{best}$ and $\Omega$. The results are below,
  \begin{center}
  \begin{tabular}{ c c c c c c }
   ratio & filters = 32    & 64    & 96   & 0 (empty layer) \\  \midrule
   $\Omega_{best}$ & 0.083  & 0.408  & 0.499 & 0.01 \\
   $\Omega$        & 0.325 & 0.325 & 0.325 & 0.025 \\
  \end{tabular}
  \end{center}
  From the table, it is apparent that networks falling in $\Omega_{best}$ tend to have more filters, which is consistent with our prior knowledge.
\end{itemize}

\end{document}